\documentclass[final,5p,times,twocolumn,authoryear]{elsarticle}

\usepackage{amssymb}
\usepackage{libertinus, libertinust1math}
\usepackage{textcomp}
\usepackage{amsmath}
\usepackage{enumitem}
\usepackage{graphicx}
\usepackage{siunitx}

\usepackage{float}
\usepackage[hidelinks]{hyperref}

\usepackage{lineno}
\usepackage{booktabs,multirow}
\usepackage{nicefrac}
\usepackage{subcaption}
\usepackage{algorithm}
\usepackage{algorithmic}
\usepackage{pifont}
\newcommand{\cmark}{\ding{51}}
\newcommand{\xmark}{\ding{55}}

\journal{Preprint}
\begin{document}

\begin{frontmatter}

\title{Deep Learning-Based Tracking and Lineage Reconstruction of Ligament Breakup}

\author[label1]{Vrushank Ahire}
\author[label2]{Vivek Kurumanghat}
\author[label1]{Mudasir Ganaie}
\author[label2]{Lipika Kabiraj\corref{cor1}}
\ead{lipika.kabiraj@iitrpr.ac.in}
\cortext[cor1]{Corresponding author.}

\affiliation[label1]{organization={Department of Computer Science and Engineering, IIT Ropar},
             city={Rupnagar},
             postcode={140001},
             state={Punjab},
             country={India}}

\affiliation[label2]{organization={Department of Mechanical Engineering, IIT Ropar},
             city={Rupnagar},
             postcode={140001},
             state={Punjab},
             country={India}}

\begin{abstract}
The disintegration of liquid sheets into ligaments and droplets involves highly transient, multi-scale dynamics that are complex to quantify from high-speed shadowgraphy images. Identification of droplets, ligaments, and blobs formed by the breakup of a liquid sheet, and their tracking from frame to frame, is an essential part of spray analysis. Analysis of such processes poses a fundamental challenge for conventional multi-object tracking frameworks, which enforce strict one-to-one temporal associations and cannot explicitly represent one-to-many fragmentation events. In this study, we present a two-stage deep learning framework for object detection (droplets and ligaments) and for the identification of temporal relationships between objects across successive frames. This framework directly models ligament deformation, fragmentation, and parent--children lineage during the disintegration of liquid sheets. In the first stage, a Faster R-CNN detector with a ResNet-50 backbone and Feature Pyramid Network is trained to identify and classify ligaments and droplets in high-speed shadowgraphy recordings of two impinging Carbopol gel jets. A morphology-preserving synthetic data generation strategy enriches the training set without introducing physically implausible configurations, achieving a held-out test $F_1$ score of up to 0.872 across fourteen systematically varied original-to-synthetic data configurations. The robustness of the model is tested on additional datasets with different atomization conditions and liquid compositions. In the second stage, a Transformer-augmented multilayer perceptron classifies inter-frame object associations into three physically motivated categories --- continuation, fragmentation (one-to-many), and non-association, using physics-informed geometric features. Despite severe class imbalance inherent to atomizing flows, the model achieves 86.1\% overall accuracy, 93.2\% precision, and perfect recall (1.00) for fragmentation events on the held-out test set and enables pairwise detection of object relationships. Together, the two stages enable preservation of parent--child lineage, and extraction of breakup statistics including fragment multiplicity and droplet size distributions. By explicitly identifying children droplets formed from fragmentation of ligaments, the framework offers an automated analysis of the primary mode of atomization.
\end{abstract}

\begin{keyword}
Fuel atomization\sep Ligament fragmentation\sep Droplet formation\sep High-speed imaging\sep Object tracking\sep Deep learning 
\end{keyword}

\end{frontmatter}


\section{Introduction}

Liquid jet and sheet atomization involves complex mechanisms of ligament formation, and fragmentation into droplets~\citep{dumouchel2008experimental, villermaux2007fragmentation}. Understanding these phenomena is crucial for numerous technical applications, including fuel injection in combustion engines, pharmaceutical atomization, agricultural spraying, and additive manufacturing~\citep{fansler2015invited, lefebvre2017atomization}. High-speed shadowgraphy and digital imaging capture the temporal evolution of ligaments and droplets with microsecond resolution~\citep{kahler2012resolution, settles2017review}. It is very challenging to obtain quantitative data from image sequences because the objects are dense, overlapping, and rapidly changing shape, which makes changes such as fragmentation and coalescence more difficult to detect.

Standard computer vision techniques for spray analysis utilize intensity thresholding, edge detection, and morphological operations to identify discrete droplets and ligaments~\citep{blaisot2005droplet, fdida2010drop}. These methods work well for separate objects under even lighting, but they struggle with overlapping areas, irregular shapes, and the differences in imaging between real experiments~\citep{chigier1996regimes, linne2013imaging}. SORT, DeepSORT, and ByteTrack are examples of classical multi-object tracking algorithms that match frames one-to-one~\citep{bewley2016simple, wojke2017simple, zhang2022bytetrack}. This fundamental constraint inhibits existing methodologies from accurately representing fragmentation occurrences, wherein a singular parent ligament separates into multiple children droplets. This makes it difficult for computers to correlate children droplets to ligaments from images captured at very short time intervals.

\begin{table*}[h]
\centering
\caption{Comparison of learning-based methods for various recently reported spray and breakup analysis. \cmark~= supported; \xmark~= not addressed.}
\label{tab:literature_comparison}
\resizebox{\textwidth}{!}{%
\begin{tabular}{llcccccl}
\toprule
Study & 
Key method & 
\shortstack{Detect \& \\ classify} & 
\shortstack{Synthetic \\ data} & 
\shortstack{Temporal \\ tracking} & 
\shortstack{1$\to$N \\ breakup} & 
\shortstack{Lineage \\ reconstruction} & 
Key limitation \\
\midrule

\citet{sibirtsev2023mask}
& Mask R-CNN 
& \cmark & \xmark & \xmark & \xmark & \xmark 
& Frame-by-frame; no temporal modeling \\

\citet{ade2024application}
& U-Net (holography) 
& \cmark & \cmark & \xmark & \xmark & \xmark 
& Detection only; no lineage modeling \\

\citet{zhang2025intelligent}
& YOLOv8 (spray ligaments) 
& \cmark & \cmark & \cmark & \xmark & \xmark 
& Tracking limited to 1-to-1 \\

\citet{jose2024machine}, \citet{jose2025ml}
& Mask R-CNN/instance segmentation 
& \cmark & \cmark & \xmark & \xmark & \xmark 
& Frame-wise statistics; no temporal learning \\

\citet{lim2024semantic}
& YOLO-based segmentation 
& \cmark & \cmark & \xmark & \xmark & \xmark 
& Segmentation-focused; no tracking \\

\citet{tretola2024machine}
& Hybrid DL + post-processing 
& \cmark & \xmark & \cmark & \xmark & \xmark 
& Breakup handled via heuristics; not learned \\

\citet{hasti2022denoising}
& Modified U-Net 
& \cmark & \xmark & \xmark & \xmark & \xmark 
& Single-frame; no ligament-level modeling \\

\citet{huynh2024real}
& YOLOv5 (droplet detection) 
& \cmark & \cmark & \xmark & \xmark & \xmark 
& Droplet-only; no ligament or breakup model \\

\midrule
Present study 
& Faster R-CNN + TransformerMLP 
& \cmark & \cmark & \cmark & \cmark & \cmark 
& \\

\bottomrule
\end{tabular}}
\end{table*}
Recent advances in deep learning techniques on images processing have transformed object detection and instance segmentation in multiphase flow applications~\citep{lecun2015deep, he2017mask}. Convolutional Neural Networks (CNNs), particularly region-based architectures like Faster R-CNN~\citep{ren2015faster} and Mask R-CNN~\citep{he2017mask}, demonstrate superior performance in detecting droplets and ligaments under various imaging conditions~\citep{sibirtsev2023mask}. Studies by \citet{jose2025ml}, \citet{jose2024machine} have shown that instance segmentation models consistently outperform traditional computer vision methods in extracting droplet size distributions and primary breakup features. Moreover, hybrid pipelines that integrate deep learning with classical post-processing techniques have demonstrated robustness across various operating conditions~\citep{tretola2024machine}. However, these approaches primarily focus on frame-by-frame detection without explicitly correlating the temporal relationships required to reconstruct fragmentation events, leaving a significant gap in the analysis of dynamic liquid sheet breakup processes.

Synthetic data creation is a useful and efficient approach for training deep learning models when available data is scarce and accurate labeling of data is tedious~\citep{ade2024application, lim2024semantic}. \citet{ade2024application} created holographic pictures to train U-Net models how to separate droplets. \citep{zhang2025intelligent} created enhanced datasets for YOLOv8 training by combining real backgrounds with labeled ligaments. These experiments show that synthetic images that keep the shape and brightness of objects can greatly improve model generalization, especially in high-speed shadowgraphy images, where annotation is a time-consuming process. However, such studies do not tracking of objects and correlate fragmented components from a parent ligament, and this is essential for comprehending fragmentation events.

Table~\ref{tab:literature_comparison} summarizes current learning-based methods reported in recent works on spray and image based analysis. Prior research focuses on detection, synthetic augmentation, and constrained 1-to-1 tracking; however, none explicitly facilitate one-to-many breakup modeling or lineage restoration. This study fills these gaps by modeling temporal linkages together and allowing for full fragmentation lineage investigation. Based on this goal, we present a two-stage deep learning architecture that integrates spatial analysis through object identification (Study I) with temporal relationship inference (Study II). The main contributions are:

\begin{itemize}[label=\tiny$\blacksquare$, leftmargin=*, itemsep=2pt, topsep=2pt, parsep=0pt, partopsep=0pt]
    \item A complete two-stage framework that combines Faster R-CNN object identification with a Transformer-augmented MLP to represent occurrences where ligaments break up.
    \item A morphology-preserving synthetic data generation strategy that achieves up to 86.9\% $F_1$ score on an unseen test set using minimal ground-truth annotations.
    \item Systematic comparison across varied original-to-synthetic data ratios, identifying optimal configurations for spray imaging.
    \item Automated extraction of fragmentation trees, parent--child lineages, and breakup statistics, with perfect recall (1.00) for infrequent breakup events.
\end{itemize}

The remainder of this paper is organized as follows. Section~\ref{sec:setup} describes the experimental setup and data acquisition. Section~\ref{sec:methodology} presents the proposed two-stage framework. Section~\ref{sec:results} reports experimental results and evaluation. Section~\ref{sec:conclusions} discusses future directions and concludes the paper.

\section{Atomization setup and data acquisition}
\label{sec:setup}
In this section, we describe the experimental setup and diagnostics used to acquire high-speed imaging data of a liquid spray. The test fluid preparation, atomization configuration, imaging setup, and annotation methodology (for training and evaluating the proposed framework) are discussed in detail.
\subsection{Test fluid preparation and rheological characterization}
A carbopol gel with a weight concentration of 0.10 wt\% is used as the test fluid. Carbopol gels are non-Newtonian fluids widely used as gel propellant simulants to examine various atomization characteristics. Our previous studies aimed to understand the mechanisms of liquid sheet breakup and droplet formation in carbopol gels in weight concentrations of 0.10--0.30 wt\%. In the present study, we chose 0.10 wt\% carbopol gel, which has the lowest zero-shear viscosity among the compositions studied, as an initial dataset. The steps we followed for preparing the gel are described in our previous articles~\citep{Saurabh2022,Kurumanghat2025}. Carbopol is a cross-linked polyacrylic acid polymer. Carbopol powder dissolves in water to form clear gels. The gel is prepared by gradually adding Carbopol powder to deionized water while stirring with a mechanical stirrer. The solution is further neutralized with triethanolamine that facilitates gelation. Rheological measurements of the 0.10~wt\% carbopol gel is performed using a
rotational rheometer equipped with a cone-and-plate geometry. Steady shear
experiments are conducted over a shear rate range of 0.001--1000~\si{\per\second}, encompassing the deformation rates relevant to the atomization process.
\begin{figure*}[h]
    \centering
    \includegraphics[width=0.95\linewidth]{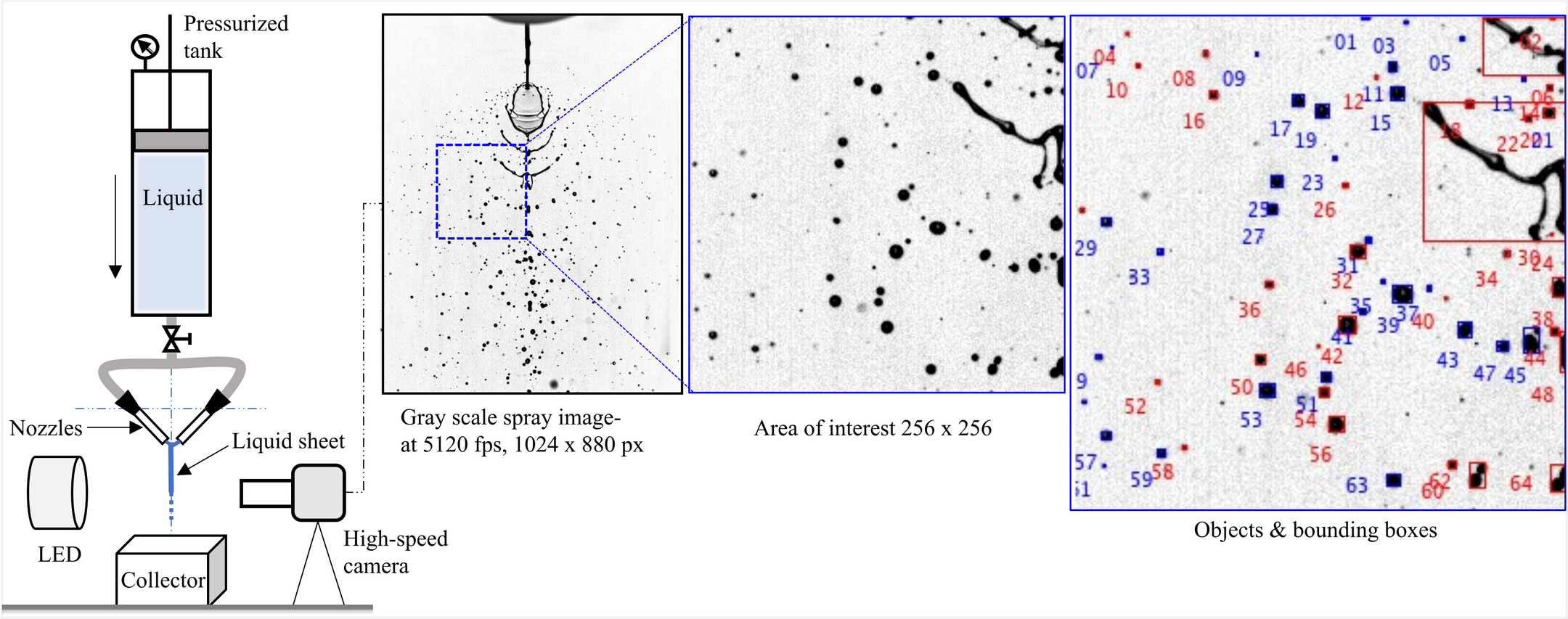}
    \caption{Experimental setup for impinging jets atomization and high-speed shadowgraphy. A pressurized piston-cylinder system drives the liquid through two impinging nozzles, forming a liquid sheet that breaks into ligaments and droplets. The imaging pipeline extracts localized $256\times256$ patches with object-level bounding box annotations (ligament~L, droplet~D). The objects are shown with indices and alternate red and blue boxes for easy visualization (best viewed in color).}
    \label{fig:experimental_setup}
\end{figure*}
\subsection{Atomization configuration}
Atomization experiments are conducted using a like-on-like doublet
impinging-jets configuration, as schematically illustrated in
Fig.~\ref{fig:experimental_setup}. The test fluid is supplied from a cylindrical reservoir
(inner diameter 72~mm) in which the liquid and the pressurizing gas are
separated by a piston. Compressed nitrogen gas is supplied to the gas side of
the reservoir through a pressure regulator, and the applied pressure displaced
the piston, thereby driving the liquid through the atomizer. The piston is
capable of traversing a distance of approximately 35--40~cm, allowing sufficient
liquid volume to be delivered under steady operating conditions.

From the reservoir, the liquid is conveyed through a
\SI[parse-numbers=false]{\nicefrac{1}{4}}{in} diameter supply line and delivered
to two identical smooth nozzles. Each nozzle had a length of \SI{1}{in}
(25.4~mm) and an inner diameter of \SI{0.41}{\milli\meter}. The jets issued from
the nozzles impinged at a fixed angle of \SI{90}{\degree}, with a pre-impingement
distance of \SI{10}{\milli\meter}, forming a thin liquid sheet at the point of
impingement. Experiments are conducted at three fixed Weber numbers of
$\mathrm{We} \approx 400$, 2780, and 9470 (see Table~\ref{tab:additional_cases}), ensuring identical inertial-to-surface-tension
conditions for the image dataset used in the machine-learning analysis.

Injection pressure is measured at the nozzle inlet using STS ATM.1ST pressure
sensors with a measurement range of 0--33~bar and a resolution of 5~kPa. The
piston displacement is measured using a resistance-based linear position
sensor mounted on the piston rod. The measured piston motion is used to
estimate the instantaneous volumetric flow rate and the corresponding jet exit
velocity. 

\subsection{High-speed shadowgraphy imaging}

High-speed imaging of the impinging jets and the resulting spray formation is
performed using a Photron Fastcam Mini AX100 camera equipped with a 105~mm macro
lens. A back-lighting configuration is employed using a continuous white LED
panel to provide uniform illumination and high contrast between the liquid
structures and the background. Images are acquired at a frame rate of
5120~frames~per~second with an image resolution of $1024 \times 880$ pixels,
corresponding to a spatial resolution of
83~\si[per-mode=symbol]{\micro\meter\per{pixel}}. All images are recorded in 8-bit grayscale
format. 

A total of 287 high-speed images are extracted from the recorded sequence under
steady operating conditions for use in the machine-learning analysis. These
images capture the liquid sheet formation, instability growth, ligament
formation, and droplet breakup processes. For temporal tracking and lineage reconstruction (Study~II), a representative short sequence of consecutive frames exhibiting pronounced ligament deformation and breakup is selected.

All sensor signals, including pressure and piston position measurements, are
acquired using a National Instruments data acquisition system equipped with a
NI~9220 voltage input module. The high-speed camera is synchronized with the
data acquisition system to ensure temporal alignment between the operating
conditions and the acquired image data. 
\begin{table*}[h]
    \centering
    \caption{Validation cases tested on robustness across Carbopol gel compositions and Weber numbers. Case 1 corresponds to the training condition.}
    \label{tab:additional_cases}
    \begin{tabular}{ccccc}
        \toprule
        Case & Carbopol Conc. (wt\%) & We & Patch & Remarks \\
        \midrule
        1 & 0.10 & 400  & Original  & Baseline case  \\
        2 & 0.10 & 400  & Different patch & Repeatability check \\
        3 & 0.20 & 2780 & Patch 1          & New concentration \\
        4 & 0.20 & 2780 & Different patch  & Repeatability check \\
        5 & 0.30 & 9470 & Patch 1          & New concentration \\
        6 & 0.30 & 9470 & Different patch  & Repeatability check \\
        \bottomrule
    \end{tabular}
\end{table*}
In the present study, we utilize high-speed images of the atomization of a 0.10 wt\% carbopol gel at a Weber number of 400 under atmospheric conditions. This composition acts as a base case to test the performance of the proposed methodology, which could be further extended to other compositions. Additional datasets listed in Table~\ref{tab:additional_cases} are used to
evaluate the trained Stage-1 detector across different
Carbopol concentrations and Weber numbers, whereas only one test condition, 0.10 wt\% Weber number 400 is used for Stage 2 of the study. The detector is applied to these cases without retraining or threshold adjustment.

\subsection{Ground truth annotation protocol}

Ground truth annotations for object detection are generated using a semi-automated computer vision pipeline developed in MATLAB. The pipeline performed initial object detection using thresholding on inverted grayscale images, followed by connected-component analysis and morphological filtering to suppress noise. The threshold is a fixed intensity value of
\textbf{150} on the 8-bit (0--255) scale, applied to the complement of the
grayscale crop, and chosen empirically to separate liquid structures from the
illuminated background across all patches. Detected objects are labeled manually as either ligaments (L) or droplets (D) based on their appearance, size and aspect ratio. The labeling is based on the geometric features of the object, such as the area-to-perimeter ratio and eccentricity. Droplets tend to be more spherical, whereas ligaments appear elongated and stretched after breakup from the sheet. 

The largest connected object in each image, corresponding to the continuous liquid sheet, is automatically identified and excluded from subsequent analysis. All detection, annotation, and tracking steps in both Study~I and Study~II are performed on cropped image patches of size $256 \times 256$ pixels extracted from the original high-resolution images. This patch-based formulation localizes regions of active breakup and ensures consistent spatial context across frames. For each detected object, geometric properties including bounding box coordinates, mask area, and centroid location are extracted. All generated annotations are manually reviewed and corrected, requiring substantially less effort than full manual annotation while ensuring high-quality ground truth. Manual annotation typically requires 5--10~min per image to label approximately 60--70 objects in Stage~1, with additional effort for Stage~2 relationship labeling. While this annotation process imposes some limitations on dataset scalability, larger, independently annotated datasets across diverse experimental conditions remain important for evaluating model convergence and generalization. Objects smaller than 4 pixels are excluded, as they approach the optical resolution limit and are highly susceptible to imaging noise.

For temporal relationship annotation (Study~II), all possible object pairs between frame $t$ and the subsequent frame $t{+}1$ within corresponding $256 \times 256$ patches are considered. Object pairs whose centroids could not be physically related are assigned the label NONE (label = $-1$) by default. This means that the relationship between any two given objects that are not related across the two frames are annotated NONE. So, every object in frame $t$, will have a matching object in frame $t{+}1$ for cases where object from frame $t$ did not break up. For break up events the matching will be one to many depending on number of children droplets formed. Explicit manual annotation is therefore restricted to physically meaningful events: MOVE (label = 0), indicating continuation of the same object across frames, and BREAKUP (label = 1), indicating fragmentation of a parent object into multiple children. This annotation method significantly reduced the amount of manual labeling needed, while still fully capturing the important temporal relationships. Additionally, the information regarding the features of objects like pixel area, centroid, bounding box intersections are also extracted. These steps are described in Section~\ref{feature_engineering}. In total, 37 consecutive frame pairs are annotated with pairwise object relationships. This forms the dataset used to train and evaluate the temporal relationship classifier.


\section{Methodology}
\label{sec:methodology}

The proposed method separates spatial recognition from temporal inference. Study~I performs object detection and classification on individual frames, whereas Study~II determines temporal relationships between detected objects across subsequent frames. Fig.~\ref{fig:two_stage_framework} illustrates a detailed two-stage architecture.

\subsection{Study I: Spatial analysis (object detection)}

The objective of Study I is to create an object detection framework capable of accurately localizing and identifying ligaments and droplets in individual frames, and thus establishing a foundation for subsequent analysis. As depicted in Fig.~\ref{fig:two_stage_framework}, the model operates on individual image patches, producing object-level predictions denoted by bounding boxes and associated class labels (ligament or droplet). During the training phase, these predictions are refined against ground truth annotations utilizing standard detection losses, enabling the model to acquire both precise localization and semantic classification abilities. The learning pipeline integrates both original and synthetic data to enhance generalization across varied data environments. Model development is conducted using a cross-validated training approach, with final performance assessed on a separate held-out set of original unseen images to ensure unbiased evaluation. The detailed methodology, encompassing data generation, model architecture, and experimental configuration, is described in the subsequent subsection.

\subsubsection{Morphology-preserving synthetic data generation}

\begin{figure}[h]
\centering
\includegraphics[width=0.47\textwidth]{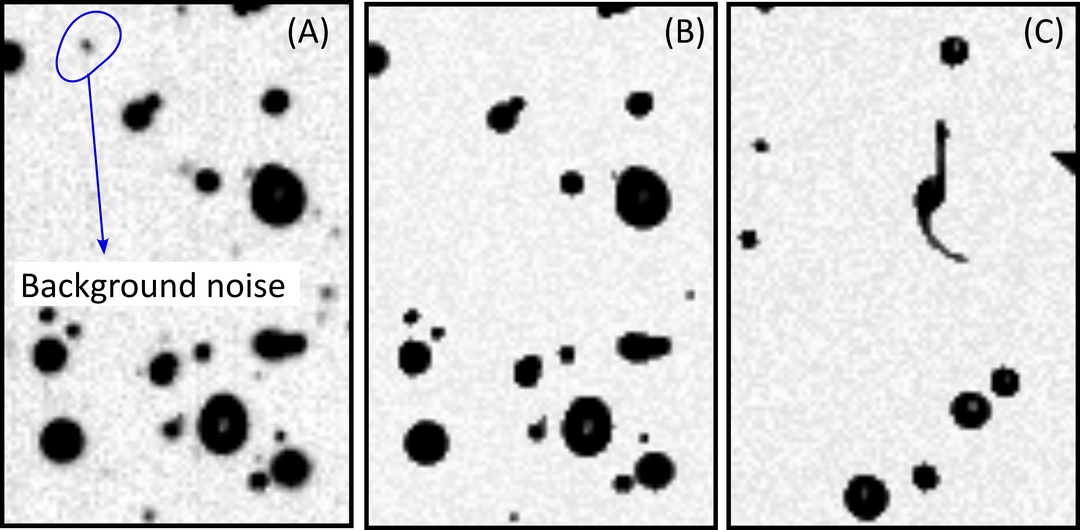}
\caption{Representative examples of morphology-preserving data generation showing (A) original image, (B) clean image, and (C) generated synthetic image.}
\label{fig:synthetic_pipeline}
\end{figure}

\begin{figure*}[h]
    \centering
    \includegraphics[width=0.98\textwidth]{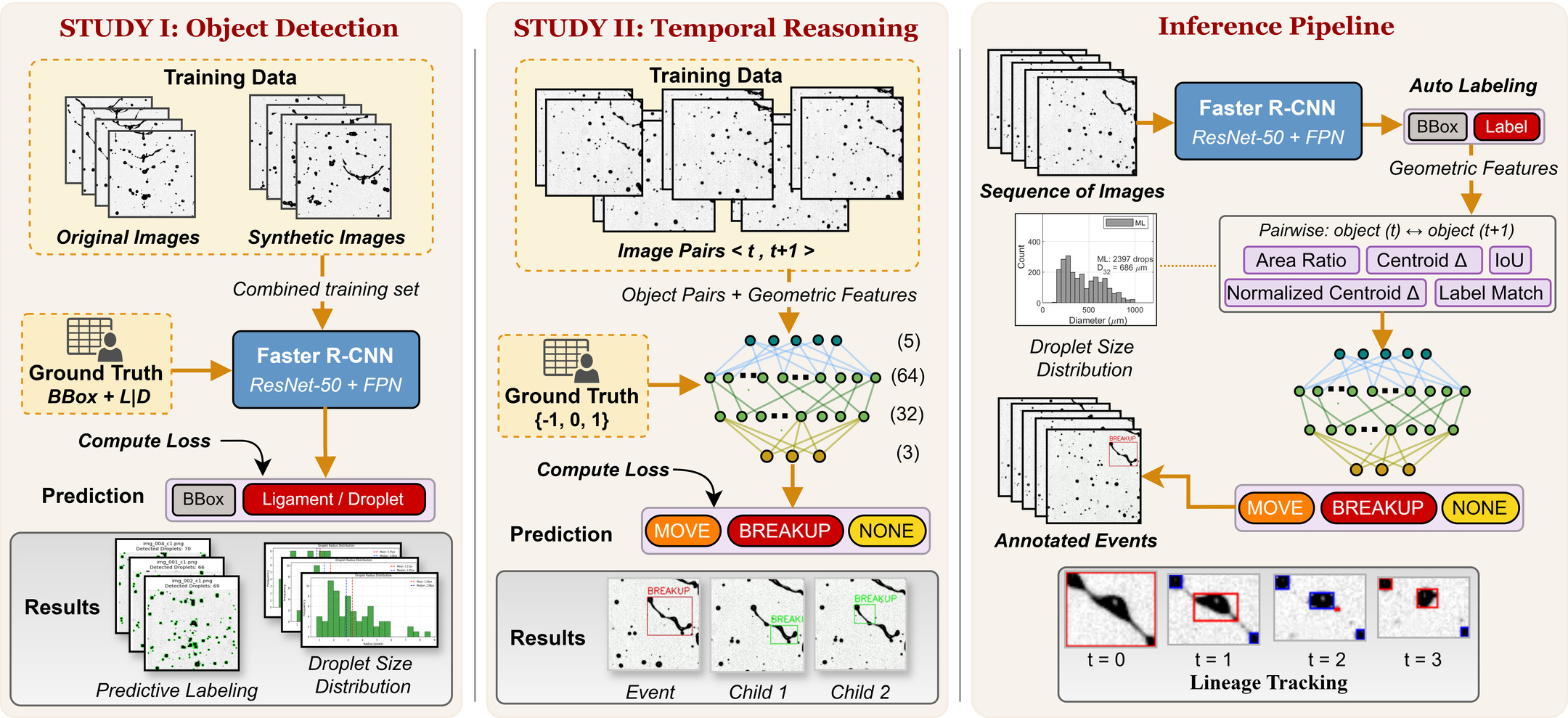}
    \caption{Proposed two-stage deep learning framework. \textbf{Study~I} performs frame-wise detection using Faster R-CNN (ResNet-50+FPN) trained on original and morphology-preserving synthetic images, outputting bounding boxes and class labels. \textbf{Study~II} encodes object pairs from consecutive frames using physics-informed geometric features and passes them to a Transformer-augmented MLP that predicts \textsc{MOVE}, \textsc{BREAKUP}, or \textsc{NONE} relationships, enabling fragmentation lineage reconstruction.}
    \label{fig:two_stage_framework}
\end{figure*}

To overcome the challenge of scarcity of labeled data, we utilize a synthetic data generation strategy. Our approach differs from conventional augmentation techniques such as rotation, flipping, or elastic deformation, which can create physically impossible forms. Instead, it preserves the original object shapes and their relative spatial arrangement as observed during breakup. Fig.~\ref{fig:synthetic_pipeline} presents illustrative examples of a patch from an \textit{original image} ($256 \times 256$ px), its corresponding \textit{clean image}, and a \textit{synthetic image}.

For each original training image, all detected ligaments and droplets are extracted along with their spatial coordinates, creating a collection of physically realistic objects. Connected-component analysis is applied to the ligament objects to preserve the largest contiguous region, eliminating small droplets in the cropped ligament image and yielding a clean ligament mask. These processed object crops are referred to as \emph{clean objects} and are tightly cropped to eliminate unnecessary background regions.

Following that, two types of training images are constructed.
Initially, \emph{clean objects} are placed onto a noise-matched background while preserving their original spatial locations. Secondly, fully \emph{synthetic composite} images are generated by randomly choosing between 50 and 70 \emph{clean objects} from the collection of ligament and droplet objects and placing them at random, non-overlapping locations on a synthetic canvas. The canvas background is generated using Gaussian noise with intensity values uniformly distributed between 225 and 255, simulating the texture of shadowgraphy backgrounds.

As all synthetic objects are drawn directly from the annotated object collection, their class labels and bounding box coordinates are predefined. As a result, no additional manual annotation are required for the synthetic images. A combination of cleaned original images and randomized composite images increases data diversity while preserving physical plausibility. This allows the detector to generalize across different regions of the original high-resolution images ($1028 \times 880$ px) and across varying object densities.

Overall, the synthetic images retained the object morphologies present in the original dataset while incorporating controlled variation in background appearance and spatial arrangement.

\subsubsection{Faster R-CNN architecture and training}

Object detection is implemented using Faster R-CNN~\citep{ren2015faster} with a ResNet-50~\citep{he2016deep} backbone and Feature Pyramid Network (FPN)~\citep{lin2017feature}. This architecture is chosen for its strong detection ability which is essential for spray images, where object sizes range from small droplets ($\sim$10 pixels) to elongated ligaments exceeding 100 pixels. The ResNet-50 backbone is initialized with weights pretrained on the COCO dataset~\citep{lin2014microsoft}, while the FPN allows feature representation across scales.

The detection steps comprises of two parallel branches for bounding box regression and object classification (background, droplet, and ligament). Region of Interest (RoI) Align~\citep{he2017mask} is used to extract feature maps from the FPN, preserving spatial alignment, which is essential for the precise localization of small objects.

Training is conducted for 25 epochs with a batch size of 4 and an initial learning rate of $5\times10^{-3}$. A step learning rate schedule is used with a decay factor of $\gamma = 0.1$ applied after every 7 epochs. Optimization is performed using stochastic gradient descent with momentum 0.9 and weight decay $5\times10^{-4}$. Gradient clipping with a maximum norm of 10.0 is used to stabilize training. Data augmentation included random horizontal flipping (probability 0.5) and photometric perturbations (brightness and contrast variations within $\pm$10\%).

During training of the model, the network optimizes the loss that includes a classification loss for object category prediction (background, droplet, and ligament) and a bounding box regression loss for precise localization. The output of the model is a set of bounding boxes with associated class labels and confidence scores, which provide accurate detection of small droplets and elongated ligaments across varying scales.

\subsubsection{Experimental configurations}
To evaluate the effect of synthetic data on detection performance by the model, three experimental configurations are designed with different proportions of number of original and synthetic training images, as described in Table~\ref{tab:training_configs}. 

\begin{table}[h!]
\centering
\caption{Training data configurations for evaluating the impact of original-to-synthetic data composition on object detection performance.}
\label{tab:training_configs}
\resizebox{\columnwidth}{!}{%
\begin{tabular}{lcccc}
\toprule
Experiment & Setting & Original & Synthetic & Total \\
\midrule
\multirow{5}{*}{Exp1} & Nog\_050\_Nsyn\_00200 (E1 a) & 50 & 200 & 250 \\
 & Nog\_100\_Nsyn\_00150 (E1 b) & 100 & 150 & 250 \\
 & Nog\_150\_Nsyn\_00100 (E1 c) & 150 & 100 & 250 \\
 & Nog\_200\_Nsyn\_00050 (E1 d) & 200 & 50 & 250 \\
 & Nog\_250\_Nsyn\_00000 (E1 e) & 250 & 0 & 250 \\
\midrule
\multirow{4}{*}{Exp2} & Nog\_250\_Nsyn\_00000 (E2 a) & 250 & 0 & 250 \\
 & Nog\_250\_Nsyn\_00250 (E2 b) & 250 & 250 & 500 \\
 & Nog\_250\_Nsyn\_00500 (E2 c) & 250 & 500 & 750 \\
 & Nog\_250\_Nsyn\_01000 (E2 d) & 250 & 1000 & 1250 \\
\midrule
\multirow{5}{*}{Exp3} & Nog\_050\_Nsyn\_00250 (E3 a) & 50 & 250 & 300 \\
 & Nog\_100\_Nsyn\_00250 (E3 b) & 100 & 250 & 350 \\
 & Nog\_150\_Nsyn\_00250 (E3 c) & 150 & 250 & 400 \\
 & Nog\_200\_Nsyn\_00250 (E3 d) & 200 & 250 & 450 \\
 & Nog\_250\_Nsyn\_00250 (E3 e) & 250 & 250 & 500 \\
\bottomrule
\end{tabular}}
\end{table}

Experiment~1 (Exp1) maintained a constant total of 250 training images while altering the original-to-synthetic image ratio across five configurations (E1a--E1e), with the number of original images increasing from 50 to 250 and synthetic images decreasing from 200 to 0. Experiment~2 (Exp2) maintained the original image count at 250 while progressively increasing the number of synthetic images across four configurations (E2a--E2d), resulting in total training sizes of 250, 500, 750, and 1250 images. Experiment~3 (Exp3) maintained a constant synthetic image count of 250 while varying the number of original images from 50 to 250 across five configurations (E3a--E3e), resulting in total training sizes ranging from 300 to 500 images.

For each configuration, the dataset is randomly shuffled and partitioned, and 5-fold cross-validation is performed on the available data to obtain training and validation splits for model development and selection. The validation data also comprises synthetic image samples to monitor performance and select the best model within each fold.

To ensure an unbiased evaluation, final performance is assessed on a separate held-out test set of 37 original unseen images, which are not used at any stage of training, validation, or synthetic data generation. The independent test set remained consistent over Exp1, Exp2, and Exp3 to enable unbiased comparison. This methodology offers a more rigorous evaluation than standard cross-validation, as it combines cross-validated model selection with a new test set for final assessment.

\subsection{Study II: Temporal relationship inference}

The aim of Study~II is to deduce physically meaningful temporal relationships between objects detected in successive frames, explicitly accounting for one-to-many fragmentation events. Rather than enforcing a one-to-one relationship as in classical multi-object tracking, we formulate temporal relations as a \emph{pairwise classification problem} over objects detected in successive frames (time instances).

\subsubsection{Pairwise formulation of temporal associations}

Consider two consecutive frames acquired at times $t$ and $t{+}1$. Let $\mathcal{O}^{(t)} = \{ o^{(t)}_1, \dots, o^{(t)}_{N_t} \}$ and $\mathcal{O}^{(t+1)} = \{ o^{(t+1)}_1, \dots, o^{(t+1)}_{N_{t+1}} \}$ denote the sets of detected objects (ligaments or droplets) in the two frames, obtained from Stage~1. Each object is associated with a bounding box, geometric attributes (described in Section~\ref{feature_engineering}), and a class label (ligament or droplet).

For every ordered object pair $(o^{(t)}_i, o^{(t+1)}_j)$, we seek to determine their semantic temporal relationship. This relationship is defined as one of three mutually exclusive classes: $y_{ij} \in \{-1, 0, 1\}$, where
\begin{itemize}[leftmargin=*, itemsep=1pt]
    \item $y_{ij} = 0$ (\textsc{MOVE}): the two objects correspond to the same physical entity across frames,
    \item $y_{ij} = 1$ (\textsc{BREAKUP}): object $o^{(t)}_i$ is the parent of object $o^{(t+1)}_j$ due to fragmentation,
    \item $y_{ij} = -1$ (\textsc{NONE}): the objects are physically unrelated.
\end{itemize}

This formulation naturally accommodates one-to-many breakup events, as a single parent object $o^{(t)}_i$ may be associated with multiple children in $\mathcal{O}^{(t+1)}$ through multiple pairwise predictions with $y_{ij}=1$.

\subsubsection{Feature engineering for physical relationship modeling}
\label{feature_engineering}

Rather than learning temporal associations directly from image appearance, each object pair $(o^{(t)}_i, o^{(t+1)}_j)$ is represented using a compact set of physically meaningful geometric features derived from object bounding boxes and binary masks.

Let $o^{(t)}_i$ and $o^{(t+1)}_j$ denote two detected objects in consecutive frames $t$ and $t{+}1$. Each object is characterized by a bounding box $\mathcal{B} = (x_1, y_1, x_2, y_2)$, centroid $\mathbf{c} = (c_x, c_y)$, binary mask $\mathcal{M}$, projected area $A$, all computed from the connected-component representation of the object mask.

For each candidate pair, a feature vector $\mathbf{f}_{ij}$ is constructed as follows:
\begin{enumerate}[label=(\arabic*), leftmargin=*]

    \item \textit{Centroid distance}, defined as 
    $\Delta c = \|\mathbf{c}_j - \mathbf{c}_i\|_2$, 
    where $\mathbf{c}_i$ and $\mathbf{c}_j$ are centroids of the object masks in frames $t$ and $t{+}1$, respectively.

    \item \textit{Normalized centroid distance}, defined as 
    $\Delta c_{\text{norm}} = \frac{\|\mathbf{c}_j - \mathbf{c}_i\|_2}{\sqrt{A_i}}$, 
    which accounts for scale-dependent displacement.

    \item \textit{Bounding box Intersection-over-Union (IoU)}, computed as 
    $\frac{|\mathcal{B}_i \cap \mathcal{B}_j|}{|\mathcal{B}_i \cup \mathcal{B}_j|}$, 
    measuring spatial overlap between the two objects.

    \item \textit{Area ratio}, given by 
    $\frac{A_j}{A_i}$, 
    where $A_i$ and $A_j$ are the pixel areas computed from the binary masks.

    \item \textit{Type consistency}, defined as a binary indicator 
    \[
    \delta_{\text{type}} = 
    \begin{cases}
    0, & \text{if } \tau_i = \tau_j \\
    1, & \text{if } \tau_i \neq \tau_j
    \end{cases},
    \]
    where $\tau \in \{\text{ligament}, \text{droplet}\}$ denotes the object class label.

\end{enumerate}

These features collectively represent essential physical characteristics associated with ligament stretching, thinning, breakup, and droplet movement. The features, such as centroid distance and normalized centroid distance, provide insight into the object's movement and how far it has traveled, while the bounding box IoU and area ratio capture the evolution of the shape of object. This facilitates distinction between object continuation (MOVE), fragmentation (BREAKUP), and objects that are unrelated (NONE) without relying on visual similarity.

The choice of a Transformer-based architecture for temporal relationship classification is motivated by the structure of the task rather than architectural preference alone. Distinguishing MOVE, BREAKUP, and NONE associations is not governed by any single geometric cue in isolation but emerges from the joint interaction of the five input features: centroid displacement, normalized displacement, bounding-box overlap, area ratio, and object-type consistency. Feature correlation analysis (Fig.~12) confirms that these inputs are largely complementary rather than redundant (max $|r| = 0.57$), indicating that the discriminative signal resides in their combined, nonlinear interplay. A self-attention mechanism is therefore well suited to this setting, as it enables the model to dynamically weight feature combinations for each object pair rather than applying fixed or additive weighting. To validate this choice empirically, five architectures of increasing relational complexity were benchmarked: BasicMLP, ResidualMLP, AttentionMLP, FeatureInteractionMLP, and TransformerMLP. The TransformerMLP consistently achieved the highest cross-validated accuracy, precision, and F1-score, together with the lowest inter-fold variance, indicating both superior and more stable generalization. On the held-out test set, it further achieved perfect recall for the rare but physically critical BREAKUP class, a result of particular importance given the severe class imbalance and the irreversible consequence of a missed fragmentation event on droplet lineage reconstruction.

To exclude physically implausible associations, spatial gating is applied by discarding object pairs with a centroid distance greater than 64 pixels or a normalized centroid distance greater than 5 pixels. These thresholds represent the maximum expected displacement in \SI{0.2}{ms}, which is the time between successive frames.

The Stage 2 dataset comprises 37 image pairs, of which 25 were used for training and validation and the remaining 12 for testing. At the object-pair level, the dataset contains 1607 MOVE pairs, 71 BREAKUP pairs, and over 8000 NONE pairs, reflecting the strong class imbalance inherent to this problem. The dominance of NONE pairs is physically expected, as the majority of object pairs across consecutive frames are spatially and temporally unrelated. Similarly, simple continuation events (MOVE) are far more frequent than fragmentation events (BREAKUP) within any given image pair sequence. Despite this imbalance, each image pair contributed one or more MOVE and BREAKUP instances, so that the limited number of image pairs still yields a meaningful sample count at the object-pair level. The TransformerMLP achieved perfect recall for the BREAKUP class on the held-out test split. Given that this result is based on 86 BREAKUP samples, it should be interpreted as strong but preliminary evidence of sensitivity to fragmentation events within the tested experimental conditions, rather than a statistically definitive performance guarantee. This limitation is acknowledged, and broader validation across a larger number of BREAKUP instances remains an important direction for future work.

\subsubsection{Neural architectures and hyperparameter optimization}

\begin{table*}[t]
\centering
\small
\caption{Neural architectures evaluated for temporal relationship classification.}
\label{tab:stage2_architectures}

\setlength{\tabcolsep}{6pt}
\renewcommand{\arraystretch}{1.15}

\begin{tabular}{l p{0.85\columnwidth}}
\toprule
Model & Architecture details \\
\midrule

BasicMLP &
$5 \rightarrow 64 \rightarrow 32 \rightarrow 3$ \\
& Linear + BN + ReLU + Dropout (0.3) \\

ResidualMLP &
$5 \rightarrow 64 \rightarrow$ ResBlock $\times 2 \rightarrow 32 \rightarrow 3$ \\
& Linear(64) + BN + ReLU + Dropout (0.2) + Linear(64) + BN \\

AttentionMLP &
$5 \rightarrow 64 \rightarrow$ Attention $\rightarrow 32 \rightarrow 3$ \\
& $\mathbf{a} = \sigma(W_2 \tanh(W_1 \mathbf{x}))$ \\

FeatureInteractionMLP &
Input (5) + interactions (10) $\rightarrow 15$ \\
& $15 \rightarrow 128 \rightarrow 64 \rightarrow 32 \rightarrow 3$ \\
& Linear + BN + ReLU + Dropout (0.1--0.3) \\

TransformerMLP &
$5 \rightarrow 64 \rightarrow$ Transformer $\rightarrow 32 \rightarrow 3$ \\
& $d_{\text{model}}=64$, FFN = 128, [CLS] token \\

\bottomrule
\end{tabular}
\end{table*}

To model temporal relationships between object pairs, we formulate a supervised multi-class classification problem using a set of five geometric features: centroid distance (pixel and normalized), bounding box Intersection over Union (IoU), area ratio, and type consistency. All features are standardized using z-score normalization prior to training.

We evaluate five neural architectures of increasing complexity. All models take a 5-dimensional input vector and output class probabilities over three categories: \textsc{MOVE}, \textsc{BREAKUP}, and \textsc{NONE}. The architectural details are summarized in Table~\ref{tab:stage2_architectures}.

All models are trained using the Adam optimizer with an initial learning rate of $10^{-3}$ and weight decay tuned to $[10^{-5}, 10^{-2}]$. Mini-batch training is performed with batch sizes of $\{32, 64, 128\}$. Early stopping with a patience of 15 epochs is used based on the validation $F_1$-score.

To address severe class imbalance (\textsc{NONE} $\gg$ \textsc{MOVE} $\gg$ \textsc{BREAKUP}), a weighted cross-entropy loss is used, where class weights are inversely proportional to class frequencies. Hyperparameter optimization, including model type, learning rate, batch size, weight decay, and spatial gating thresholds, is executed using Bayesian optimization (Optuna) with a Tree-structured Parzen Estimator (TPE) sampler. Model performance is evaluated using 5-fold cross-validation, and the final model is selected based on the highest validation $F_1$-score.

\subsection{Lineage reconstruction and end-to-end inference pipeline}

The final stage of the framework integrates the outputs of Study~I (spatial detection) and Study~II (temporal relationship modeling) to reconstruct complete object lineages across time. This end-to-end inference pipeline is illustrated in Fig.~\ref{fig:two_stage_framework}.

Given a sequence of images, Study~I is first applied independently to each frame to detect ligaments and droplets, along with their corresponding bounding boxes. For each pair of consecutive frames $(t, t{+}1)$, all spatially gated object pairs $(o^{(t)}_i, o^{(t+1)}_j)$ are then generated and encoded using the geometric feature vectors described in Section~\ref{feature_engineering}. These feature vectors are passed to the trained classifier from Study~II, which provides class probabilities for the three temporal relationship types as outputs: \textsc{MOVE}, \textsc{BREAKUP}, and \textsc{NONE}.

The predicted pairwise relationships represent physically consistent temporal associations. For each object $o^{(t)}_i$ in frame $t$, candidate children objects in frame $t{+}1$ with predicted probabilities exceeding predefined thresholds ($\tau_{\text{move}} = 0.5$ for \textsc{MOVE} and $\tau_{\text{break}} = 0.3$ for \textsc{BREAKUP}) are identified. If only a single \textsc{MOVE} association is present and no competing breakup predictions, a continuation edge is assigned that preserves object identity across frames.

When several objects exceed the \textsc{BREAKUP} threshold for the same parent object, the event is termed as fragmentation. In such cases, the subset of children objects that minimizes the area conservation error
$
\left| A_i - \sum_k A_{j_k} \right| / A_i
$
is selected, enforcing approximate mass conservation and ensuring physical plausibility. All remaining associations are rejected and categorized as \textsc{NONE}.

The resultant temporal associations are organized into a directed acyclic graph (DAG), where nodes represent detected objects at specific time instances, and edges denote continuation or fragmentation events. This graph topology directly represents parent--children relationships and supports one-to-many breakup events. The graph traversal facilitates automated reconstruction of fragmentation trees, from which breakup statistics such as fragment multiplicity, ligament lifespan, parent-to-child area ratios, and object velocities (derived from centroid displacements) can be directly estimated.

\section{Results and discussion}
\label{sec:results}
We present the quantitative and qualitative evaluation of the proposed two-stage framework. The results are presented for Study I and Study II as introduced in the methodology. Through Study~I, we demonstrate the ability of a trained Faster R-CNN detector to locate and classify ligaments and droplets. The results describe how different original--synthetic data compositions influence detection accuracy, geometric consistency, and droplet-size statistics. Subsequently, through Study~II we demonstrate the temporal relationship classifier that classifies \textsc{MOVE}, \textsc{BREAKUP}, and \textsc{NONE} associations between consecutive frames. This further enables breakup-aware tracking and ligament lineage reconstruction. These results collectively demonstrate that the proposed framework provides both reliable object detection and temporal relationships for fragmentation in sprays.

\subsection{Study I: Object detection performance}

The Faster R-CNN detector is evaluated on a held-out benchmark that comprises 37 original unseen images. Tables~\ref{tab:exp1_results}--\ref{tab:exp3_results} summarize the test-set precision, recall, and $F_1$-score for the three experiments--Exp1, Exp2, and Exp3, whereas Table~\ref{tab:crossval_results} reports the corresponding 5-fold cross-validation results in terms of mean $\pm$ standard deviation. In addition to standard detection metrics, average precision is reported for droplets (D) and ligaments (L), which allows a detailed assessment of class-specific performance. Figs.~\ref{fig:scatter_exp1}--\ref{fig:hist_exp3} visualize the geometric consistency and droplet-size distributions obtained from the computer-vision (CV) annotations and the model predictions. Fig.~\ref{fig:stage1_qualitative} shows qualitative results on two different configurations.

\begin{figure*}[h]
\centering
\includegraphics[width=0.98\textwidth]{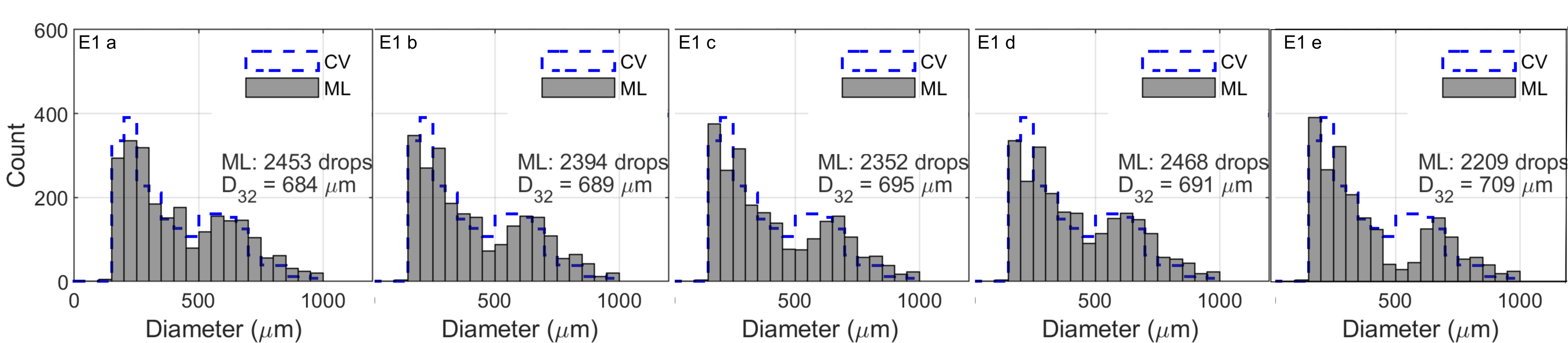}
\caption{Droplet diameter histograms for Experiment~1 comparing CV-derived and model-predicted distributions across five training configurations. E1 (a--e) shows five experiment setting as discussed in Table~\ref{tab:training_configs}. Mixed training settings better preserve the overall shape of the droplet-size distribution.}
\label{fig:hist_exp1}
\end{figure*}

\begin{figure*}[h]
\centering
\includegraphics[width=0.98\textwidth]{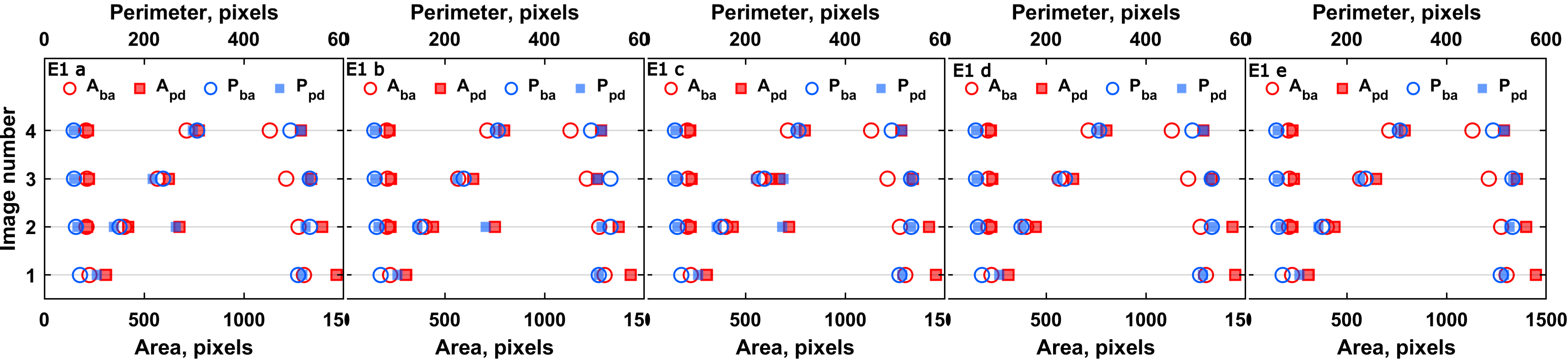}
\caption{Scatter plots for Experiment~1 comparing predicted and ground-truth geometric features across five training configurations. E1 (a--e) shows five experiment setting as discussed in Table~\ref{tab:training_configs}. Mixed original--synthetic settings exhibit closer agreement with the reference annotations than the all-original baseline. A and P denote area and perimeter, respectively, where \textit{ba} and \textit{pd} denote \textit{base} and \textit{predicted}, respectively.}
\label{fig:scatter_exp1}
\end{figure*}

The results indicate that the incorporation of synthetic data is most effective when used to complement rather than replace original experimental images. Across the three experiments, moderate synthetic augmentation improves generalization, particularly when labeled data availability is less, whereas excessive reliance on synthetic data results in reduced localization fidelity or class imbalance. The results also reveal a precision--recall trade-off: configurations with more original images tend to produce higher precision, whereas mixed original--synthetic settings often improve recall and object sensitivity. This trade-off is especially relevant for spray analysis, where missed droplets or unresolved neighboring objects can directly affect downstream droplet-size statistics and breakup characterization.

\subsubsection{Experiment 1: Effect of original-to-synthetic ratio}

Here we discuss the results obtained from training RCNN model using image sets as shown in Table~\ref{tab:exp1_results}. The variation in the ratio of original to synthetic images affects precision, recall and $F_1$ score. For a fixed number (250) of images, when comparing test sets that use only original images during training, Detailed results are discussed here. It is clear that incorporating synthetic data improves held-out test performance. It is observed that for the case ExpI set 4, original images to synthetic images in the ration 200/50 shows the highest Precision, Recall and F-1 score.

\begin{table}[h!]
\centering
\caption{Object detection performance for Experiment~1 (fixed total = 250 images, varying original/synthetic ratio). Evaluated on 37 held-out original images. \textbf{Bold} indicates the best value per column.}
\label{tab:exp1_results}
\resizebox{\columnwidth}{!}{%
\begin{tabular}{lccc}
\toprule
Setting & Precision & Recall & $F_1$-score \\
\midrule
Nog\_050\_Nsyn\_00200 & 0.836 & 0.859 & 0.847 \\
Nog\_100\_Nsyn\_00150 & 0.832 & 0.856 & 0.844 \\
Nog\_150\_Nsyn\_00100 & 0.813 & 0.836 & 0.825 \\
Nog\_200\_Nsyn\_00050 & \textbf{0.827} & \textbf{0.879} & \textbf{0.852} \\
Nog\_250\_Nsyn\_00000 & 0.760 & 0.778 & 0.769 \\
\bottomrule
\end{tabular}}
\end{table}

The original set (250 original, 0 synthetic) attains the lowest $F_1$-score of 0.769. All of the mixed setups perform better, with improvements of up to 8.3\%. The best held-out performance comes from \texttt{Nog\_200\_Nsyn\_00050}, which gets an $F_1$-score of 0.852 and the highest recall of 0.879. This means that a training set with primarily actual images and a small number of synthetic samples is a good way to get a mix of reality and variety. Notably, even the very synthetic configuration \texttt{Nog\_050\_Nsyn\_00200} gets an $F_1$ score of 0.847, which shows that the images that are made still have enough morphological information to help detection performance.

We examine how well the model performs for each subgroup in Exp1 by comparing the droplet size distributions from model predictions with those from manually labeled ground truth annotations. Fig.~\ref{fig:hist_exp1} shows histogram plots of the droplet size distribution for each of the five parts of Exp1. The number of droplets detected using CV is 2347, and the corresponding D$_{32}$ = 617~\si{\micro\meter}. The Sauter Mean Diameter (SMD) from the model predictions is shown in each subplot. The model-detected droplet count differs from the manual count across all data subsets. The SMD data further supports this finding. They show not only variations in detection counts but also potential biases in the droplet-size distribution.

Fig.~\ref{fig:scatter_exp1} shows how well the ligaments are found. In this case, ligaments are defined by their predicted area and perimeter. Findings from four typical test photos are displayed for each Exp1 subgroup. This comparison shows how well the model distinguishes between ligament morphology and its manually determined counterparts. It shows changes in shape and scale that may not be fully captured by aggregate statistics alone. The scatter graphs in Fig.~\ref{fig:scatter_exp1} show that the predicted geometric descriptors match the actual trends, especially for the mixed-data setups. The histogram comparisons in Fig.~\ref{fig:hist_exp1} show that the predicted droplet-size distributions are still in line with the CV reference distributions. 

This experiment suggests that a mixture of original images thus enhances the count and accuracy of object detection. Similar analyses are done for Exp2 and Exp3, and the droplet size distribution histograms and ligament characterization plots for each are presented in Figs.~\ref{fig:hist_exp2}--\ref{fig:scatter_exp2} and Figs.~\ref{fig:hist_exp3}--\ref{fig:scatter_exp3}, respectively. The tendencies observed in these tests are examined in the next sections, focusing on how well the model performs overall and how synthetic image augmentation affects it.

\subsubsection{Experiment 2: Scaling synthetic data with fixed original images}

\begin{table}[h!]
\centering
\caption{Object detection performance for Experiment~2 (fixed 250 original images, increasing synthetic data). Evaluated on 37 held-out original images. \textbf{Bold} indicates the best value per column.}
\label{tab:exp2_results}
\resizebox{\columnwidth}{!}{%
\begin{tabular}{lccc}
\toprule
Setting & Precision & Recall & $F_1$-score \\
\midrule
Nog\_250\_Nsyn\_00000 & 0.764 & 0.776 & 0.770 \\
Nog\_250\_Nsyn\_00250 & 0.837 & 0.880 & \textbf{0.858} \\
Nog\_250\_Nsyn\_00500 & \textbf{0.847} & 0.864 & 0.856 \\
Nog\_250\_Nsyn\_01000 & 0.847 & \textbf{0.850} & 0.849 \\
\bottomrule
\end{tabular}}
\end{table}

\begin{figure*}[h]
\centering
\includegraphics[width=0.98\textwidth]{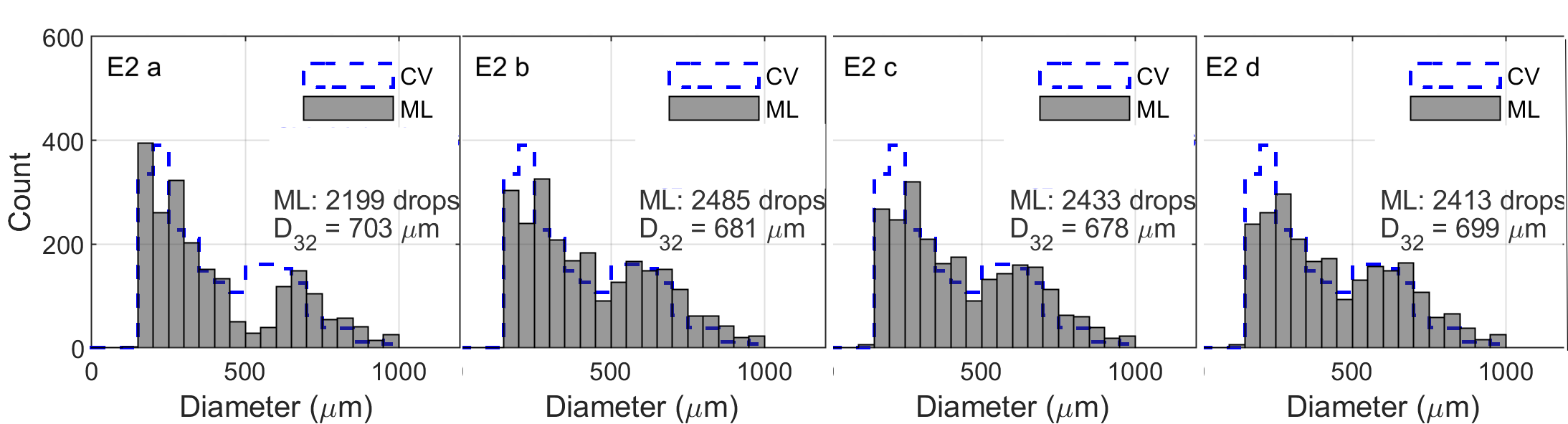}
\caption{Droplet diameter histograms for Experiment~2. E2 (a--d) shows four experiment setting as discussed in Table~\ref{tab:training_configs}. Increasing synthetic data improves agreement between CV and model-predicted droplet-size distributions up to a moderate augmentation level.}
\label{fig:hist_exp2}
\end{figure*}

\begin{figure*}[h]
\centering
\includegraphics[width=0.98\textwidth]{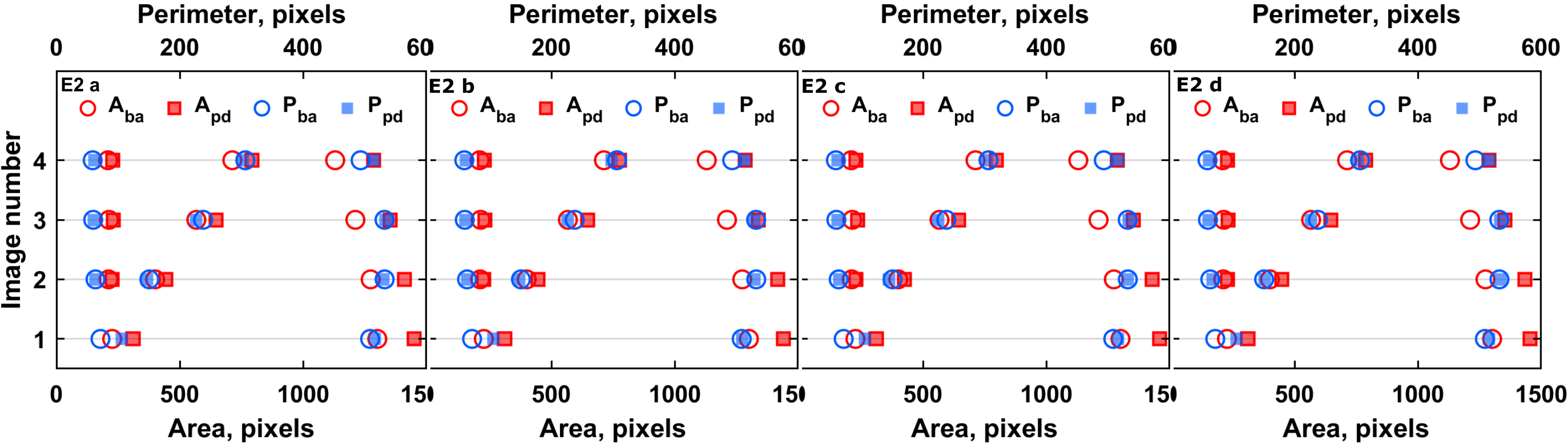}
\caption{Scatter plots for Experiment~2 illustrating the relationship between predicted and ground-truth geometric features as the synthetic data volume increases. E2 (a--d) shows four experiment setting as discussed in Table~\ref{tab:training_configs}. Agreement improves with moderate augmentation, while the gains become marginal at higher synthetic ratios. A and P denote area and perimeter, respectively, where \textit{ba} and \textit{pd} denote \textit{base} and \textit{predicted}, respectively.}
\label{fig:scatter_exp2}
\end{figure*}

Experiment~2 analyzes the effect of increasing the amount of synthetic data while maintaining the number of original image counts at 250. The results are summarized in Table~\ref{tab:exp2_results}.

A considerable improvement is observed when synthetic data is integrated into to a fixed collection of 250 original images. The $F_1$-score improves from 0.770 for the no-synthetic baseline to 0.858 for \texttt{Nog\_250\_Nsyn\_00250}. Increasing the synthetic count to 500 maintains strong performance ($F_1$ = 0.856), but further increasing it to 1000 results in a slight decline ($F_1$ = 0.849). These results indicate that moderate augmentation is beneficial, but that very large synthetic-to-original ratios yield low returns and may bias the detector toward synthetic-specific appearance patterns. 

In practical terms, the most optimal augmentation range in this experimental context is 250--500 synthetic images, corresponding to approximately one to two times synthetic expansion relative to the original dataset. The corresponding histogram and scatter plots in Figs.~\ref{fig:hist_exp2} and~\ref{fig:scatter_exp2} support these findings. The predictions align with the reference geometric measurements as synthetic data is introduced, but saturate at higher augmentation levels. Synthetic images improve robustness and coverage of plausible morphologies, but excessive augmentation does not yield proportional gains on the test-set.

\subsubsection{Experiment 3: Effect of original data volume with fixed synthetic images}

Experiment~3 studies the effect of increasing the number of original images while keeping the synthetic data budget fixed at 250 images. The results are provided in Table~\ref{tab:exp3_results}.

\begin{table}[h!]
\centering
\caption{Object detection performance for Experiment~3 (fixed 250 synthetic images, varying original data volume). E3 (a--e) shows five experiment setting as discussed in Table~\ref{tab:training_configs}. Evaluated on 37 held-out original images. \textbf{Bold} indicates the best value per column.}
\label{tab:exp3_results}
\resizebox{\columnwidth}{!}{%
\begin{tabular}{lccc}
\toprule
Setting & Precision & Recall & $F_1$-score \\
\midrule
Nog\_050\_Nsyn\_00250 & \textbf{0.862} & 0.864 & 0.863 \\
Nog\_100\_Nsyn\_00250 & 0.848 & 0.861 & 0.855 \\
Nog\_150\_Nsyn\_00250 & 0.835 & 0.853 & 0.844 \\
Nog\_200\_Nsyn\_00250 & 0.849 & 0.877 & 0.863 \\
Nog\_250\_Nsyn\_00250 & 0.861 & \textbf{0.882} & \textbf{0.872} \\
\bottomrule
\end{tabular}}
\end{table}

\begin{figure*}[h]
\centering
\includegraphics[width=0.98\textwidth]{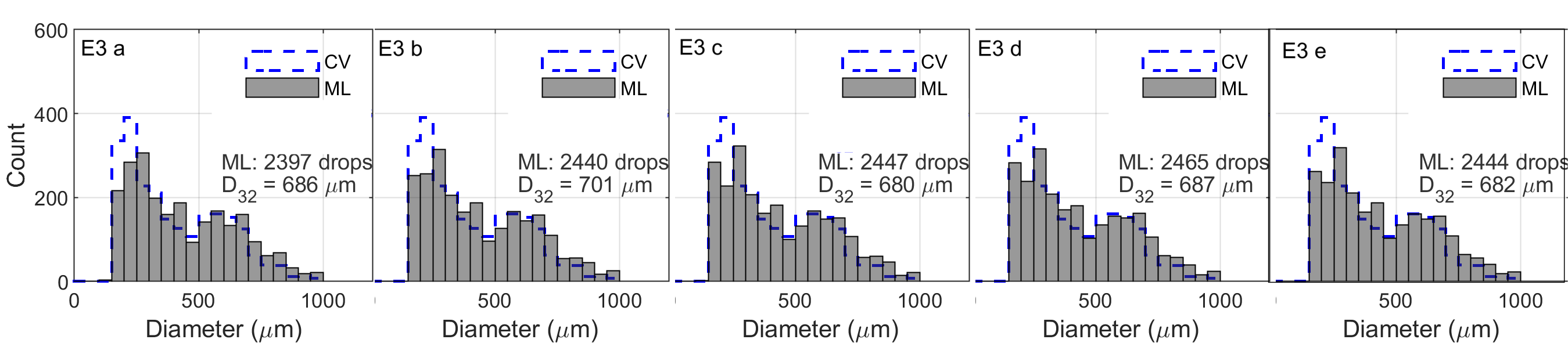}
\caption{Droplet diameter histograms for Experiment~3 showing strong agreement between CV-derived and model-predicted droplet-size distributions across all settings, with the best agreement observed at higher original data volumes. E3 (a--e) shows five experiment setting as discussed in Table~\ref{tab:training_configs}.}
\label{fig:hist_exp3}
\end{figure*}

\begin{figure*}[h]
\centering
\includegraphics[width=0.98\textwidth]{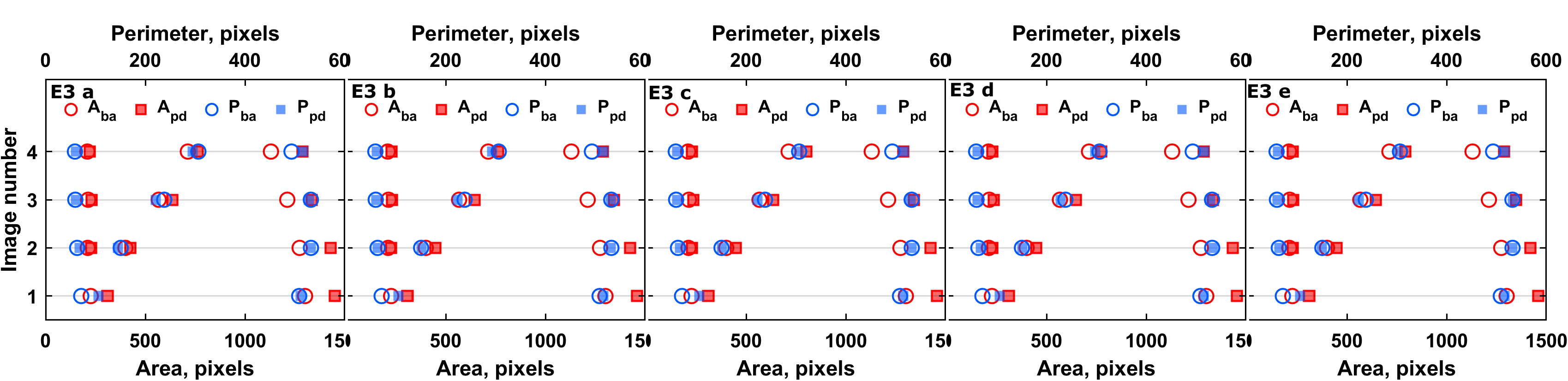}
\caption{Scatter plots for Experiment~3 illustrating the effect of increasing original data volume while keeping the synthetic set fixed. Agreement with ground-truth geometry improves as more original images are included. E3 (a--e) shows five experiment setting as discussed in Table~\ref{tab:training_configs}. A and P denote area and perimeter, respectively, where \textit{ba} and \textit{pd} denote \textit{base} and \textit{predicted}, respectively.}
\label{fig:scatter_exp3}
\end{figure*}

Results from Experiment~3 show that increasing the fraction of original images in the training data sets consistently improves generalization when paired with a fixed amount of synthetic augmentation. The highest overall held-out performance in Study~I is achieved by \texttt{Nog\_250\_Nsyn\_00250}, which attains precision = 0.861, recall = 0.882, and $F_1$ = 0.872. This outcome indicates that synthetic data remains valuable even when a moderate number of original images are considered, but its effect is maximum when sufficient original examples are also included. The configuration with only 50 original images and 250 synthetic images still achieves $F_1$ = 0.863, highlighting the effectiveness of synthetic augmentation in scenarios where limited annotated datasets are available.

The trend observed in Fig.~\ref{fig:scatter_exp3} further suggests that larger original datasets enhance geometric consistency between predictions and reference annotations. Similarly, the histograms in Fig.~\ref{fig:hist_exp3} show increasingly accurate reconstruction of droplet-size distributions as the fraction of original training data grows. These observations indicate that synthetic data enhances the feature space, whereas original images provide accurate boundary localization and separation of neighboring objects.

\subsubsection{5-Fold cross-validation results}

\begin{table*}[t]
\centering
\caption{5-fold cross-validation results for all experimental configurations in Study~I. Values are reported as mean $\pm$ standard deviation. D: droplet, L: ligament. \textbf{Bold} indicates the best mean value within each experimental block for the primary metrics mAP50, mAP50--95, precision, recall, and $F_1$-score.}
\label{tab:crossval_results}
\resizebox{\textwidth}{!}{%
\begin{tabular}{llccccccccc}
\toprule
Exp & Setting & mAP50 & mAP50--95 & Precision & Recall & $F_1$-score & AP50\_D & AP50\_L & AP50--95\_D & AP50--95\_L \\
\midrule
\multirow{5}{*}{Exp1}
& Nog\_050\_Nsyn\_00200 & 0.763 $\pm$ 0.007 & 0.469 $\pm$ 0.006 & 0.670 $\pm$ 0.006 & 0.763 $\pm$ 0.010 & 0.713 $\pm$ 0.008 & 0.697 $\pm$ 0.012 & 0.830 $\pm$ 0.008 & 0.272 $\pm$ 0.010 & 0.666 $\pm$ 0.010 \\
& Nog\_100\_Nsyn\_00150 & 0.728 $\pm$ 0.017 & 0.450 $\pm$ 0.014 & 0.749 $\pm$ 0.006 & 0.790 $\pm$ 0.008 & 0.769 $\pm$ 0.006 & 0.743 $\pm$ 0.011 & 0.713 $\pm$ 0.027 & 0.358 $\pm$ 0.008 & 0.542 $\pm$ 0.022 \\
& Nog\_150\_Nsyn\_00100 & 0.763 $\pm$ 0.008 & 0.511 $\pm$ 0.003 & 0.801 $\pm$ 0.003 & 0.795 $\pm$ 0.008 & 0.798 $\pm$ 0.005 & 0.743 $\pm$ 0.010 & 0.783 $\pm$ 0.010 & 0.406 $\pm$ 0.004 & 0.616 $\pm$ 0.008 \\
& Nog\_200\_Nsyn\_00050 & \textbf{0.787 $\pm$ 0.008} & 0.546 $\pm$ 0.006 & 0.845 $\pm$ 0.001 & \textbf{0.831 $\pm$ 0.014} & 0.838 $\pm$ 0.007 & 0.786 $\pm$ 0.018 & 0.789 $\pm$ 0.008 & 0.478 $\pm$ 0.009 & 0.614 $\pm$ 0.006 \\
& Nog\_250\_Nsyn\_00000 & 0.776 $\pm$ 0.010 & \textbf{0.580 $\pm$ 0.007} & \textbf{0.936 $\pm$ 0.002} & 0.812 $\pm$ 0.015 & \textbf{0.869 $\pm$ 0.009} & \textbf{0.801 $\pm$ 0.016} & 0.752 $\pm$ 0.008 & \textbf{0.546 $\pm$ 0.011} & 0.614 $\pm$ 0.005 \\
\midrule
\multirow{4}{*}{Exp2}
& Nog\_250\_Nsyn\_00000 & 0.807 $\pm$ 0.007 & \textbf{0.619 $\pm$ 0.007} & \textbf{0.937 $\pm$ 0.003} & 0.811 $\pm$ 0.009 & \textbf{0.869 $\pm$ 0.006} & 0.798 $\pm$ 0.011 & 0.817 $\pm$ 0.010 & \textbf{0.543 $\pm$ 0.008} & 0.694 $\pm$ 0.011 \\
& Nog\_250\_Nsyn\_00250 & 0.791 $\pm$ 0.006 & 0.523 $\pm$ 0.005 & 0.805 $\pm$ 0.003 & \textbf{0.844 $\pm$ 0.003} & 0.824 $\pm$ 0.003 & \textbf{0.814 $\pm$ 0.004} & 0.767 $\pm$ 0.013 & 0.445 $\pm$ 0.003 & 0.601 $\pm$ 0.010 \\
& Nog\_250\_Nsyn\_00500 & \textbf{0.818 $\pm$ 0.002} & 0.538 $\pm$ 0.003 & 0.774 $\pm$ 0.005 & 0.816 $\pm$ 0.004 & 0.795 $\pm$ 0.004 & 0.776 $\pm$ 0.005 & 0.860 $\pm$ 0.005 & 0.387 $\pm$ 0.008 & 0.690 $\pm$ 0.005 \\
& Nog\_250\_Nsyn\_01000 & 0.795 $\pm$ 0.005 & 0.513 $\pm$ 0.003 & 0.757 $\pm$ 0.005 & 0.761 $\pm$ 0.007 & 0.759 $\pm$ 0.005 & 0.713 $\pm$ 0.008 & \textbf{0.877 $\pm$ 0.004} & 0.323 $\pm$ 0.003 & \textbf{0.704 $\pm$ 0.004} \\
\midrule
\multirow{5}{*}{Exp3}
& Nog\_050\_Nsyn\_00250 & 0.794 $\pm$ 0.003 & 0.494 $\pm$ 0.003 & 0.690 $\pm$ 0.006 & 0.761 $\pm$ 0.004 & 0.724 $\pm$ 0.004 & 0.696 $\pm$ 0.005 & \textbf{0.891 $\pm$ 0.004} & 0.284 $\pm$ 0.004 & \textbf{0.705 $\pm$ 0.003} \\
& Nog\_100\_Nsyn\_00250 & 0.770 $\pm$ 0.002 & 0.476 $\pm$ 0.003 & 0.698 $\pm$ 0.003 & 0.776 $\pm$ 0.004 & 0.735 $\pm$ 0.003 & 0.720 $\pm$ 0.005 & 0.820 $\pm$ 0.006 & 0.309 $\pm$ 0.004 & 0.642 $\pm$ 0.004 \\
& Nog\_150\_Nsyn\_00250 & 0.779 $\pm$ 0.005 & 0.490 $\pm$ 0.003 & 0.738 $\pm$ 0.003 & 0.812 $\pm$ 0.004 & 0.773 $\pm$ 0.003 & 0.761 $\pm$ 0.004 & 0.796 $\pm$ 0.010 & 0.369 $\pm$ 0.002 & 0.612 $\pm$ 0.009 \\
& Nog\_200\_Nsyn\_00250 & 0.809 $\pm$ 0.001 & 0.535 $\pm$ 0.001 & 0.765 $\pm$ 0.003 & 0.825 $\pm$ 0.003 & 0.793 $\pm$ 0.003 & 0.775 $\pm$ 0.005 & 0.843 $\pm$ 0.003 & 0.392 $\pm$ 0.004 & 0.679 $\pm$ 0.003 \\
& Nog\_250\_Nsyn\_00250 & \textbf{0.817 $\pm$ 0.005} & \textbf{0.557 $\pm$ 0.003} & \textbf{0.798 $\pm$ 0.003} & \textbf{0.842 $\pm$ 0.002} & \textbf{0.820 $\pm$ 0.002} & \textbf{0.802 $\pm$ 0.004} & 0.831 $\pm$ 0.009 & \textbf{0.441 $\pm$ 0.004} & 0.673 $\pm$ 0.005 \\
\bottomrule
\end{tabular}}
\end{table*}

The cross-validation findings in Table~\ref{tab:crossval_results} validates that the detector performance is consistent across folds, with generally low standard deviations for all metrics. For most configurations, the variation remains below 0.015, indicating that the observed performance differences are not caused by unstable training behavior. Here, mAP50 denotes mean Average Precision at an IoU threshold of 0.50, whereas mAP50--95 calculates the mean Average Precision across IoU thresholds ranging from 0.50 to 0.95, thereby offering a thorough assessment of localization quality. Precision measures the fraction of predicted detections that are correct, recall measures the fraction of ground-truth objects that are successfully detected, the $F_1$-score summarizes their balance through the harmonic mean, and the class-wise AP values report detection performance separately for droplets (D) and ligaments (L).

The cross-validation method evaluates performance based on the fold-wise test split defined within each experimental configuration, while model selection is performed using the corresponding fold-wise validation split. Consequently, the cross-validation results reflect performance based on the data configuration used when constructing each fold, rather than the static held-out benchmark of 37 unseen original images used in Tables~\ref{tab:exp1_results}--\ref{tab:exp3_results}. Thus, the absolute rankings of configurations in cross-validation do not precisely reflect the patterns observed in the held-out benchmark, and the two evaluations should be interpreted as complementary rather than interchangeable.

In Experiment~1, the mixed configuration \texttt{Nog\_200\_Nsyn\_00050} the highest cross-validated mAP50 (0.787) and recall (0.831), whereas the original setting \texttt{Nog\_250\_Nsyn\_00000} gives the best mAP50--95 (0.580), precision (0.936), and $F_1$-score (0.869). This indicates that a modest amount of synthetic augmentation can improve object coverage and detection quality. Whereas the set with only original images remains more conservative and yields better precision and stricter IoU localization performance.

In Experiment~2, the no-synthetic baseline \texttt{Nog\_250\_Nsyn\_00000} yields the highest cross-validated mAP50--95, precision, and $F_1$-score, while \texttt{Nog\_250\_Nsyn\_00250} attains the highest recall, and \texttt{Nog\_250\_Nsyn\_00500} gives the highest mAP50. This pattern shows that moderate synthetic augmentation improves sensitivity and IoU = 0.50 detection coverage, but larger synthetic ratios do not translate into improved fine localization across stricter IoU thresholds. The reduction in $F_1$-score at \texttt{Nog\_250\_Nsyn\_01000} further suggests that excessive synthetic augmentation may reduce the consistency of the detector.

In Experiment~3, the trends exhibit greater coherence and are favorable for mixed training. As the number of original images increases while keeping 250 synthetic images fixed, all primary cross-validation metrics improve steadily, resulting in the best overall configuration \texttt{Nog\_250\_Nsyn\_00250}. This setting attains the highest mAP50 (0.817), mAP50--95 (0.557), precision (0.798), recall (0.842), and $F_1$-score (0.820) in Experiment~3. These findings indicate that synthetic images are most beneficial when they supplement a sufficiently strong base of original data sets rather than replace them.

The class-wise AP values gives further insight into detector performance. Ligament detection outperforms droplet detection, particularly at stricter IoU thresholds, as reflected by the generally higher AP50--95 values for class L. Such behavior is anticipated because ligaments are larger, more continuous, and visually easier to localize than small, closely spaced droplets. Conversely, droplet localization remains challenging, especially under stricter overlap requirements, which explains the comparatively lower AP50--95\_D values across many test settings. These observations are further supported with the qualitative examples in Fig.~\ref{fig:stage1_qualitative}, where missed or merged droplets remain the dominant failure mode.

The cross-validation results support the main conclusion of Study I: synthetic data are advantageous, however the effectiveness is largely influenced by their integration with actual images. Moderate synthetic augmentation enhances recall and detection coverage, but superior localization fidelity and overall robustness are attained when synthetic images are combined with an adequately large collection of original experimental sets.

\begin{figure*}[h!]
\centering
\includegraphics[width=0.98\linewidth]{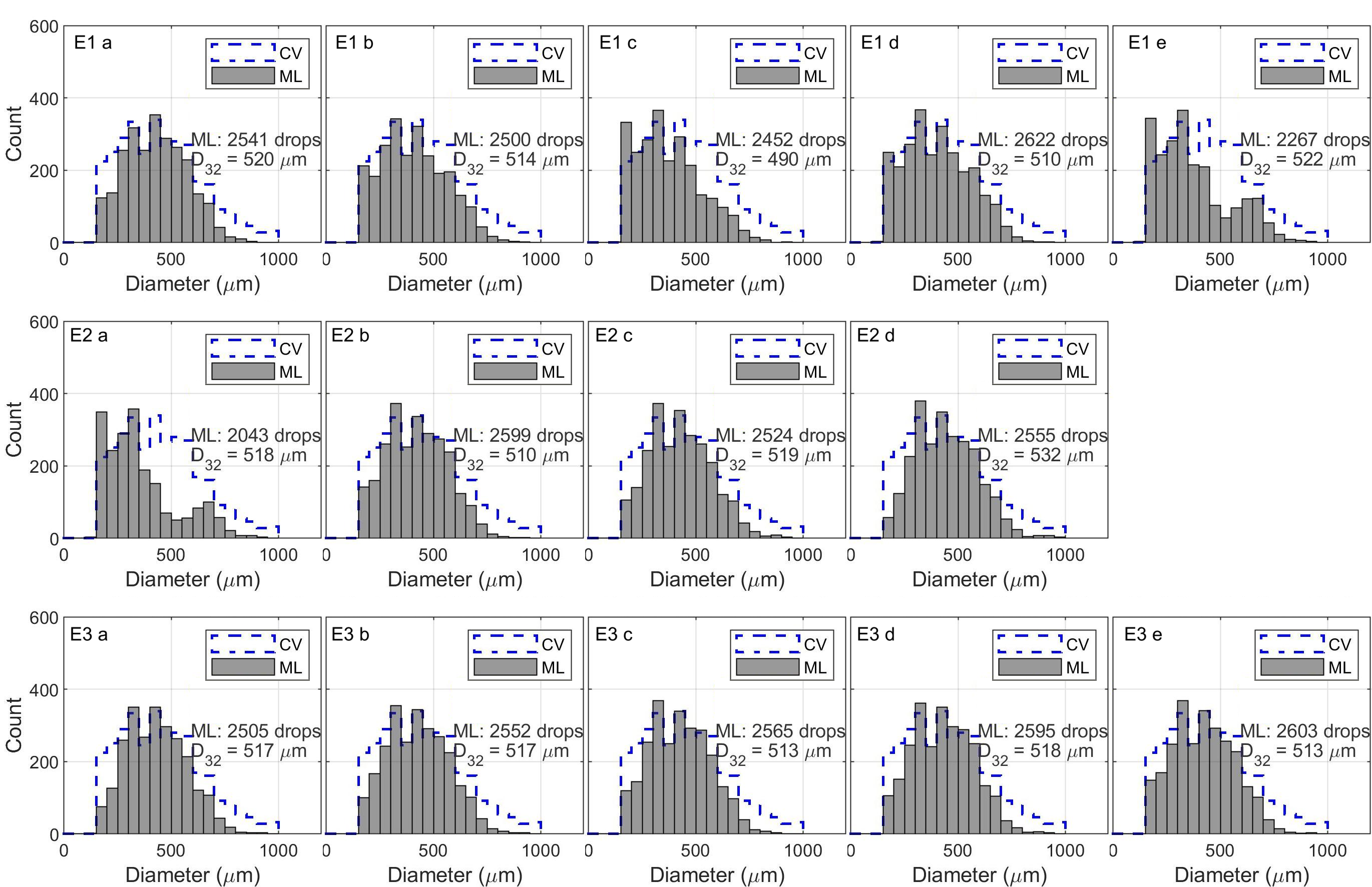}
\caption{Droplet diameter histograms for 0.20 wt\% Carbopol gel. Comparison between CV-derived and model-predicted droplet-size distributions across all three experiment sets is shown. Exp1, Exp2, and Exp3 are shown in rows. E1 (a--e), E2 (a--d), and E3 (a--e) hows experiment setting as discussed in Table~\ref{tab:training_configs}.}
\label{fig:hist_c20}
\end{figure*}

Fig.~\ref{fig:hist_c20} shows the comparison (model-predicted and computer vision-derived) of droplet size distribution of 0.20 wt\% Carbopol gel at We 2780. Results for all three experiments are mentioned in Table~\ref{tab:training_configs}. Row-1 demonstrates comparison of Computer Vision-based results with model-predicted results (Exp1). Similarly, row-2 and row-3 show comparison for Exp2 and Exp3. 

\subsubsection{Robustness across gel composition and Weber number}
\label{sec:robustness}

The model performance across the different experimental configurations is shown, including the corresponding F1-score surface plot for better visualization.
 
\begin{figure*}[h]
    \centering
    \includegraphics[width=0.8\textwidth]{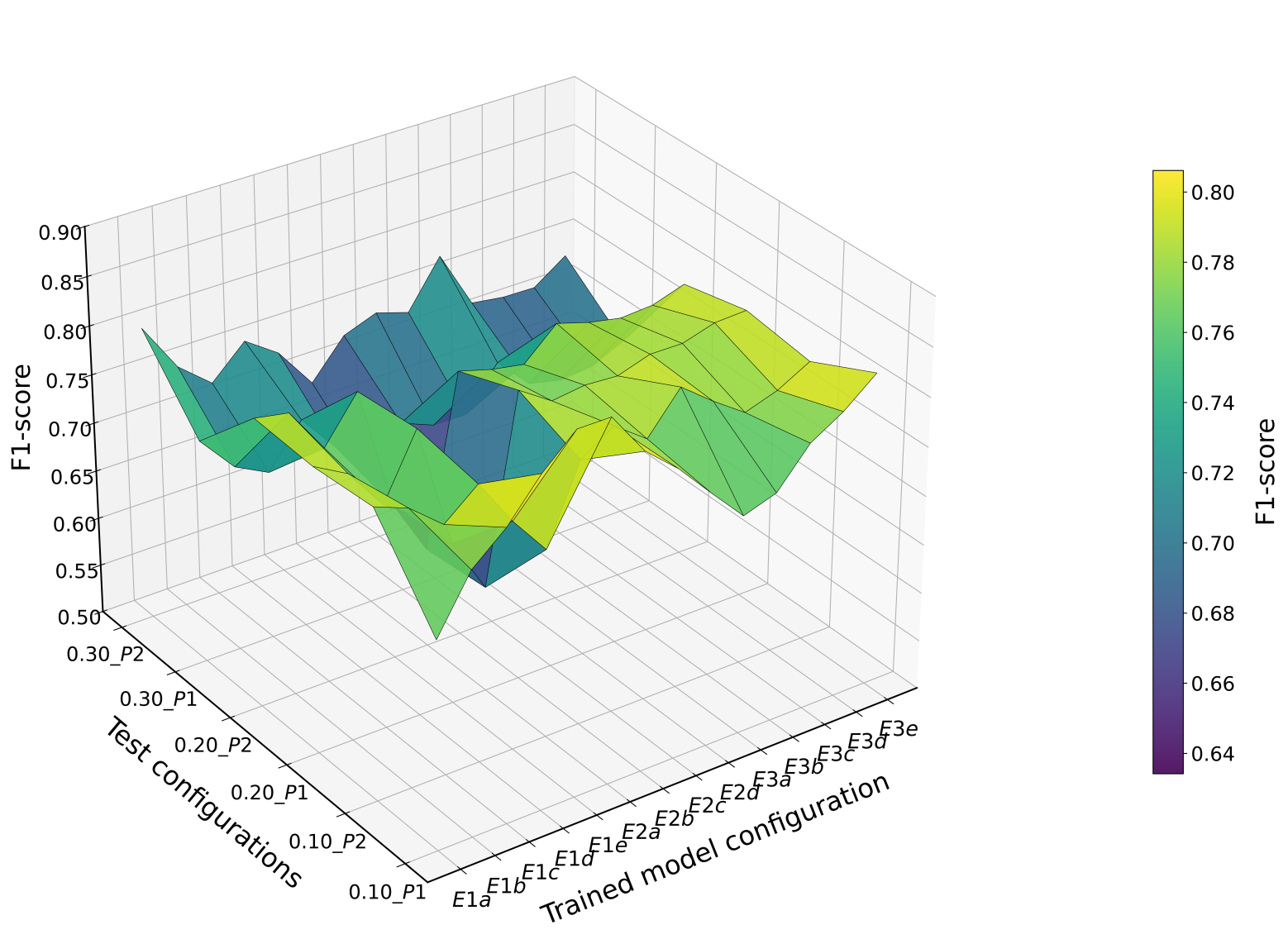}
    \caption{Cross-condition performance of the Stage-1 object detector, represented by F1-scores for 14 training configurations evaluated across six testing conditions, with the complete results provided in the Supplementary Sheet. The testing conditions comprise three Carbopol concentrations (0.10, 0.20, and 0.30~wt\%) and their corresponding Weber numbers, with two image patches (P1 and P2) evaluated for each condition. The 14 training configurations comprise five configurations from E1 (E1a--e), four from E2 (E2a--d), and five from E3 (E3a--e), illustrating the detector's generalization across different training-data compositions and experimental conditions.}
    \label{fig:f1_surface}
\end{figure*}

The model was trained on the original training data and subsequently tested on the different experimental configurations to assess its ability to generalize across changes in experimental conditions. The surface plot, Fig.~\ref{fig:f1_surface}, shows the F1-score as a function of the model performance on test configurations. Across the tested configurations, the F1-score generally ranges from approximately 0.60 to 0.84, demonstrating that the model maintains useful detection and relationship-classification performance across the different conditions. The results also show systematic differences between the testing conditions: the 0.10~wt\% cases generally yield higher F1-scores, whereas the 0.30~wt\% cases tend to be more challenging, with the 0.20~wt\% cases generally falling between these two regimes. The two patches within each experimental condition also show comparable performance, indicating robustness of the model over the effect of local image characteristics.

These results provide an additional assessment of model generalization under changes in Carbopol concentration and Weber number. In particular, the model continues to identify objects and their pairwise relationships when the fragment population becomes denser, smaller, and less spherical. The model retains approximately 85\% of the objects identified by the CV pipeline at 0.20 and 0.30~wt\%. Across the additional cases, the model also produces systematically smaller $D_{32}$ values than the CV method: 628--672~\si{\micro\meter} versus 824~\si{\micro\meter} at 0.10~wt\%, 398--455~\si{\micro\meter} versus 667~\si{\micro\meter} at 0.20~wt\%, and 475--521~\si{\micro\meter} versus 766~\si{\micro\meter} at 0.30~wt\%. The largest difference occurs at 0.20~wt\% and $\mathrm{We}=2780$, where the higher fragment density leads to increased merging of connected objects in the CV pipeline. Conversely, for the 0.30~wt\% patch~2 condition, the model gives 554--590~\si{\micro\meter}, compared with 583~\si{\micro\meter} from CV, a difference of approximately 5\%. Thus, these experiments primarily demonstrate the generalization of the trained model across different experimental conditions.

\subsubsection{Influence of synthetic images}

\begin{figure*}[h!]
\centering
\includegraphics[width=0.98\linewidth]{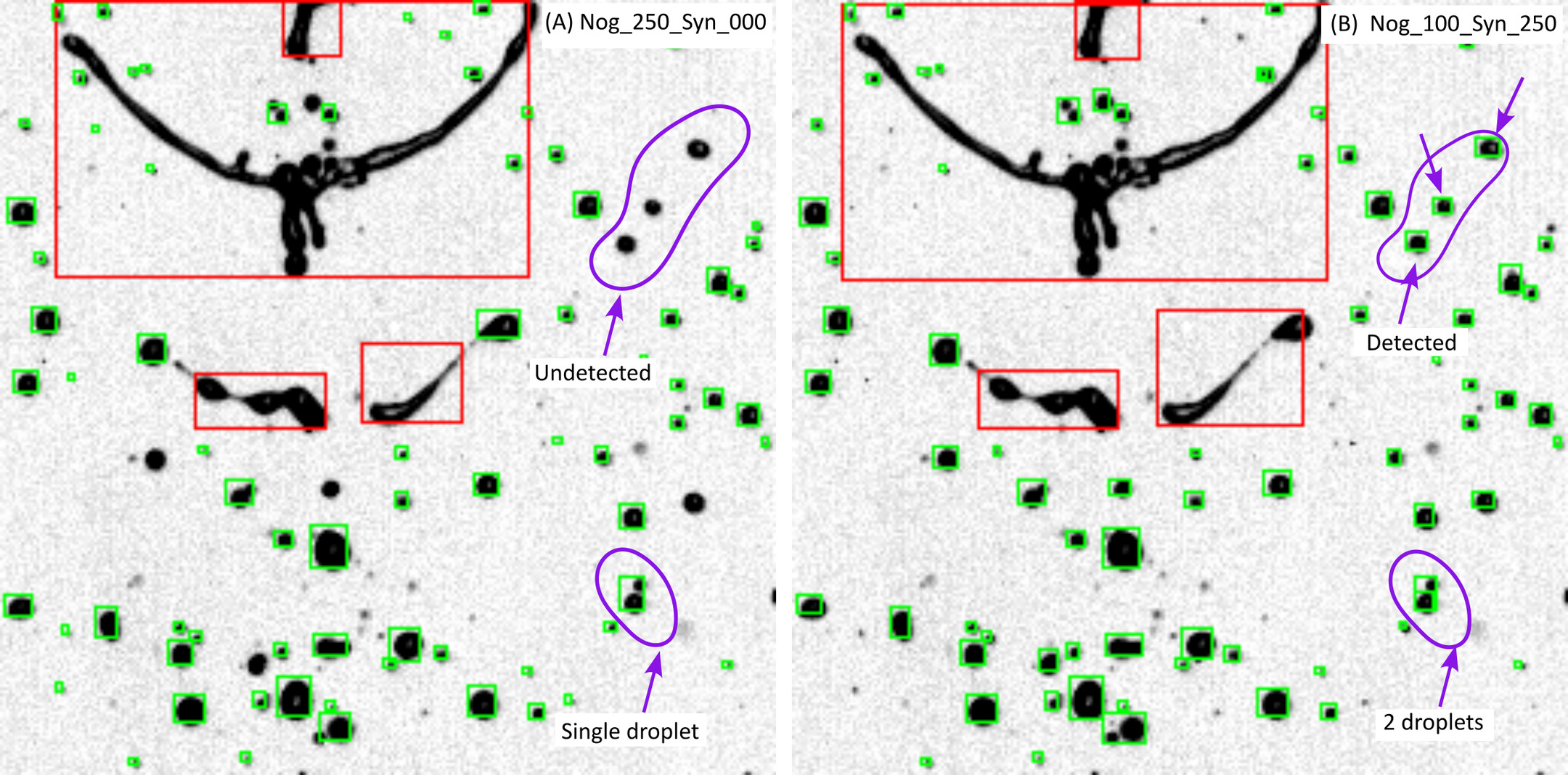}
\caption{Comparison of object detection and classification results for models trained with different data compositions. Ligaments and droplets are denoted by red and green bounding boxes, respectively. (A) Detections produced by a model trained only on original images (\texttt{Nog\_250\_Syn\_000}). (B) Detections produced by a model trained on a mixed original--synthetic dataset (\texttt{Nog\_100\_Syn\_250}). Purple-highlighted regions indicate representative failure and recovery cases.}
\label{fig:stage1_qualitative}
\end{figure*}

Fig.~\ref{fig:stage1_qualitative} illustrates the impact of training data composition on detection performance by comparing object detection and classification outputs from two models trained with varying original--synthetic data compositions. Ligaments are indicated by red bounding boxes, while droplets are denoted by green bounding boxes. Fig.~\ref{fig:stage1_qualitative}(A) presents detections from a model exclusively trained on original images (\texttt{Nog\_250\_Syn\_000}), while Fig.~\ref{fig:stage1_qualitative}(B) displays detections from a model trained on a hybrid dataset comprising both original and synthetic images (\texttt{Nog\_100\_Syn\_250}).

Two failure cases of the original-only model are indicated by purple curves in Fig.~\ref{fig:stage1_qualitative}(A). In the first highlighted region, multiple droplets are found to be missed. In the second, a pair of closely spaced droplets of different sizes is merged into a single detected object. These errors indicate that training only on original images may lack the necessary variability to effectively capture all fine droplet configurations, especially for small and neighboring droplets that warrant precise boundary marking.

In contrast, the mixed-data model depicted in Fig.~\ref{fig:stage1_qualitative}(B) effectively identifies the previously undetected droplets and distinguishes the connected droplet pair into two unique entities. This behavior aligns with the quantitative findings: augmenting original images with synthetic samples enhances sensitivity to complex structures and assists the detector in more effectively resolving fine-scale objects. Accurate droplet separation significantly influences droplet-size distributions and breakage statistics; thus, these qualitative enhancements are not solely appealing but important to the physical validity of the comprehensive spray-analysis process.

\subsubsection{Influence of objects falling on the boundary}
Partially cropped objects may introduce a potential source of bias and that their effect should be explicitly evaluated rather than assumed to be negligible.

The ROI in our study was defined according to the availability of ligament and droplet breakup regions. A raw image can contain several stages of the breakup process, including liquid-sheet breakup, ligament separation, and subsequent droplet formation. Since sheet-breakup analysis is outside the scope of the present study, the ROI was intentionally restricted to localities containing ligament breakup and the subsequent droplet/ligament structures of interest. Given the limited number of available experimental images, we initially retained all information within these ROIs.

We performed additional controlled experiments using the same experimental image sets, in which the boundary-intersecting objects were discarded by masking them out of the images. These were compared against the original model. We considered three conditions: (A) the original model, trained and evaluated with all objects including those intersecting the ROI boundary, (B) masking of the boundary-intersecting objects with the Faster R-CNN model trained from scratch, and (C) masking followed by fine-tuning of the original detector.
 
\begin{itemize}[leftmargin=*]
    \item \textbf{Case (A) Original model (baseline):} All objects within the ROI, including those intersecting the ROI boundary, are retained during training and evaluation. These are the results reported for the original model.
    \item \textbf{Case (B) Object masking without fine-tuning:} The boundary-intersecting objects are masked, and the Faster R-CNN model is trained from scratch.
    \item \textbf{Case (C) Object masking with fine-tuning:} The same masking procedure is applied, followed by fine-tuning of the Faster R-CNN model that was already trained on the original dataset.
\end{itemize}

The results are shown in Table~\ref{tab:boundary_ablation}. Masking the boundary-intersecting objects does not produce a large change in detection performance relative to the original model. For the 50\% original-data/50\% synthetic-data setting \texttt{Nog\_050\_Nsyn\_00200}, the original model achieved an mAP@0.5 of $0.763 \pm 0.007$ and an F1-score of $0.713 \pm 0.008$. Masking without fine-tuning resulted in an mAP@0.5 of $0.7925 \pm 0.0154$ and an F1-score of $0.7409 \pm 0.0068$. After fine-tuning the original model using the masked data, the corresponding values were $0.7939 \pm 0.0080$ and $0.7352 \pm 0.0107$, respectively. For the \texttt{Nog\_100\_Nsyn\_00150} setting, the original model achieved an mAP@0.5 of $0.728 \pm 0.017$ and an F1-score of $0.769 \pm 0.006$. The corresponding values were $0.7087 \pm 0.0115$ and $0.7651 \pm 0.0050$ without fine-tuning, compared with $0.7035 \pm 0.0075$ and $0.7493 \pm 0.0059$ after fine-tuning.

Thus, masking the boundary-intersecting objects changed the detection performance only moderately, and not in a consistent direction. The $\text{mAP}_{0.5}$ increased by approximately 0.03 for \texttt{Nog\_050\_Nsyn\_00200} and decreased by approximately 0.02--0.03 for \texttt{Nog\_100\_Nsyn\_00150}, and the F1-scores remained within approximately 0.03 of the original model. The results therefore indicate that the reported performance is not predominantly attributable to the inclusion of such objects.

\begin{table*}[h]
    \centering
    \caption{Ablation study with/without objects intersecting the ROI boundary.}
    \label{tab:boundary_ablation}
    \resizebox{0.8\textwidth}{!}{%
    \begin{tabular}{llccc}
        \toprule
        Setting & Metric & \begin{tabular}[c]{@{}c@{}}Case A:\\ Original model\end{tabular} & \begin{tabular}[c]{@{}c@{}}Case B:\\ Masking, no fine-tuning\end{tabular} & \begin{tabular}[c]{@{}c@{}}Case C:\\ Masking $+$ fine-tuning\end{tabular} \\
        \midrule
        \texttt{Nog\_050\_Nsyn\_00200} & mAP@0.5 & $0.763 \pm 0.007$ & $0.7925 \pm 0.0154$ & $0.7939 \pm 0.0080$ \\
                                       & F1      & $0.713 \pm 0.008$ & $0.7409 \pm 0.0068$ & $0.7352 \pm 0.0107$ \\
        \midrule
        \texttt{Nog\_100\_Nsyn\_00150} & mAP@0.5 & $0.728 \pm 0.017$ & $0.7087 \pm 0.0115$ & $0.7035 \pm 0.0075$ \\
                                       & F1      & $0.769 \pm 0.006$ & $0.7651 \pm 0.0050$ & $0.7493 \pm 0.0059$ \\
        \bottomrule
    \end{tabular}}
\end{table*}

Test images for Case A (original model) contain all objects, whereas test images for Cases B and C contain only those objects that do not lie on the boundary, with the boundary objects completely masked during both training and testing; therefore, the boundary objects do not play any role in the evaluation of Cases B and C.
 
\paragraph{Study I summary}
Taken together, the held-out test, cross-validation analysis, scatter plots, histogram comparisons, and qualitative examples show that synthetic augmentation is most effective when combined with original experimental images. Synthetic data substantially reduces the dependence on large manually annotated datasets and enables strong performance even in low-annotation regimes. However, consistent results are obtained when synthetic images supplement, rather than replace, original images.

\subsection{Study II: Temporal relationship classification}

Study~II investigates how precisely the proposed framework identifies temporal relationships between detected objects in the spray across consecutive frames. The task is to classify each candidate pair as \textsc{MOVE}, \textsc{BREAKUP}, or \textsc{NONE}, thereby facilitating breakup-aware tracking and lineage reconstruction. The subsequent section compares multiple classifier architectures, evaluates the held-out performance of the best-performing model, and examines how the chosen geometric features help in reliable temporal association.

\subsubsection{Model comparison}

Table~\ref{tab:model_comparison} compares the five evaluated architectures for temporal relationship classification, with results averaged over the cross-validation folds. The comparison shows a clear and consistent advantage for the TransformerMLP across all reported metrics.

\begin{table*}[t]
\centering
\caption{Model comparison for temporal relationship classification (Study~II). \textbf{Bold} indicates the best value.}
\label{tab:model_comparison}
\begin{tabular}{lccccc}
\toprule
Model & Avg Acc. & Avg Prec. & Avg Rec. & Avg $F_1$ $\pm$ Std & Overall $F_1$ \\
\midrule
\textbf{TransformerMLP} & \textbf{0.876} & \textbf{0.936} & \textbf{0.876} & \textbf{0.898 $\pm$ 0.006} & \textbf{0.898} \\
FeatureInteractionMLP & 0.841 & 0.921 & 0.841 & 0.867 $\pm$ 0.009 & 0.867 \\
AttentionMLP & 0.838 & 0.921 & 0.838 & 0.864 $\pm$ 0.011 & 0.864 \\
ResidualMLP & 0.834 & 0.920 & 0.834 & 0.861 $\pm$ 0.012 & 0.861 \\
BasicMLP & 0.806 & 0.909 & 0.806 & 0.836 $\pm$ 0.012 & 0.836 \\
\bottomrule
\end{tabular}
\end{table*}

Among all evaluated architectures, TransformerMLP demonstrates the strongest performance, with an average $F_1$-score of $0.898 \pm 0.006$, the highest average accuracy (0.876), precision (0.936), recall (0.876), and the lowest inter-fold standard deviation. The combination of higher mean performance and minimal variance signifies that the model is not only more precise but also more consistent across various train-validation splits. The enhancement over the BasicMLP baseline is significant, elevating the overall $F_1$-score from 0.836 to 0.898, thereby affirming that the temporal relationship classification challenge is better addressed by architectures that can model complex feature interactions instead of depending exclusively on shallow or purely direct decision boundaries.

This performance pattern aligns with the fundamental framework of the task. The distinction between \textsc{MOVE}, \textsc{BREAKUP}, and \textsc{NONE} is determined not by any single geometric indicator, but by the combined dynamics of displacement, overlap, and area variation. The TransformerMLP first projects the five-dimensional input vector into a higher-dimensional latent space, enhanced by a learnable classification token, and then processes it through stacked transformer encoder layers before final classification. This formulation, while not independently tokenizing each feature like certain tabular transformer variations, offers a more expressive method for representing nonlinear dependencies in the learnt feature representation compared to traditional feedforward baselines. The improved expressive capacity presumably enhances the efficacy of the model in addressing unclear temporal relationships.

The moderate performance of FeatureInteractionMLP, AttentionMLP, and ResidualMLP further strengthens this interpretation. All three surpass the BasicMLP, suggesting that improved feature integration, interaction modeling, or optimization stability is advantageous for this issue. Nevertheless, none equals the comprehensive precision and resilience of TransformerMLP. This indicates that partial interaction models alone is inadequate for covering the complete relational complexities of breakup dynamics. Consequently, TransformerMLP is designated as the definitive model for Study II and is employed in the ensuing qualitative and held-out test analyses.

\begin{figure*}[h!]
\centering
\includegraphics[width=0.9\linewidth]{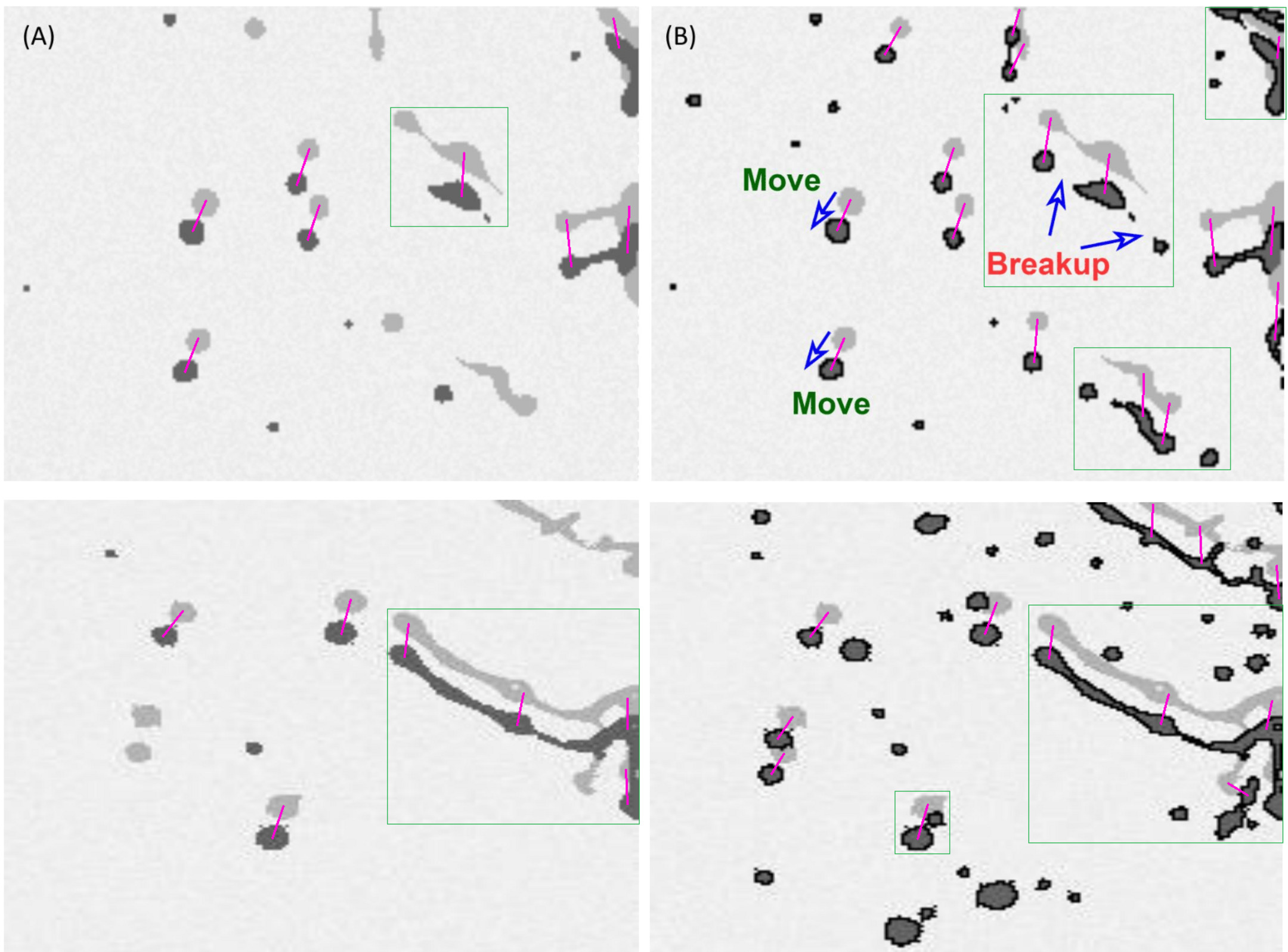}
\caption{Visualization of ligament and droplet tracking at a downstream location following initial sheet breakup. (A) Ground-truth correlations between two time instances separated by $\Delta t = 0.20$ ms, obtained via manual labeling. Lighter-shaded objects represent structures at $t=0$, while darker-shaded objects represent the corresponding structures at $t=0.20$ ms. (B) Model-predicted correlations for the same two time instances. Lighter-shaded objects correspond to $t=0$, while darker-shaded objects with prominent borders denote structures identified and correlated by the trained model at $t=0.20$ ms. Each row represents different pairs of images.}
\label{fig:stage2_qualitative}
\end{figure*}

To qualitatively evaluate temporal tracking ability, Fig.~\ref{fig:stage2_qualitative} compares manually established correspondences with model-predicted correspondences for two frames separated by $\Delta t = 0.20$ ms at a fixed downstream location after the initial sheet breakup. In Fig.~\ref{fig:stage2_qualitative}(A), the lighter and darker shades denote structures at $t=0$ and $t=0.20$ ms, respectively, with pairwise links determined through manual annotation. These annotations provide the qualitative reference against which the model predictions are evaluated.

Fig.~\ref{fig:stage2_qualitative}(B) illustrates the predictions generated by the trained TransformerMLP-based pipeline. The visual coherence between the two panels indicates that the model successfully retrieves most of the manually defined correspondences, encompassing instances of both translational displacement and geometric changes due to breakdown. That setting is complex due to the potential for objects to migrate, distort, shrink, vanish, or divide into several progeny within a brief temporal span. The qualitative examples thus corroborate the quantitative findings by demonstrating that the model generates temporal relationships instead of merely optimizing aggregate scalar measures.

\subsubsection{Temporal relationship classifier performance}

Table~\ref{tab:final_performance} reports the final evaluation of the TransformerMLP on the held-out test split. The held-out test results confirm that the performance advantages observed during cross-validation translate effectively to unseen data. TransformerMLP achieves an overall accuracy of 0.861, precision of 0.932, recall of 0.861, and $F_1$-score of 0.887, demonstrating strong generalization despite the ambiguity and severe imbalance (NONE:MOVE:BREAKUP $\approx$ 117:22:1) inherent to temporal relationship classification in atomizing flows.

\begin{table}[h!]
\centering
\caption{Final test results for the TransformerMLP model on the held-out test split in Study~II.}
\label{tab:final_performance}
\begin{tabular}{lc}
\toprule
Metric/Configuration & Value \\
\midrule
\multicolumn{2}{l}{\textit{Overall Performance}} \\
Accuracy & 0.861 \\
Precision & \textbf{0.932} \\
Recall & 0.861 \\
$F_1$-score & \textbf{0.887} \\
\midrule
\multicolumn{2}{l}{\textit{Class-wise Recall}} \\
MOVE & 0.895 \\
BREAKUP & \textbf{1.000} \\
NONE & 0.853 \\
\midrule
\multicolumn{2}{l}{\textit{Spatial Gating Constraints}} \\
Distance threshold (pixel) & 64~px \\
Distance threshold (normalized) & 8.0 \\
\bottomrule
\end{tabular}
\end{table}

From the standpoint of downstream lineage reconstruction, the most critical result is the perfect BREAKUP recall of 1.000. This means that every true fragmentation event in the held-out test split is successfully identified by the model. Such behavior is essential for breakup-aware tracking, because a missed breakup event cannot be recovered later and would permanently sever the parent--child structure of the fragmentation tree. In this setting, high breakup sensitivity is therefore more important than aggressively minimizing all positive breakup predictions, since implausible candidates can still be filtered during downstream lineage reconstruction using physically motivated constraints such as area-conservation-based consistency checks.

The remaining specific to class recall metrics explain the functional performance of the model. The MOVE recall of 0.895 implies that the model retains the majority of valid temporal continuations, which is crucial for sustaining track continuity between successive frames. The NONE recall of 0.853 indicates that the classifier effectively dismisses most invalid associations, thereby minimizing the introduction of spurious edges into the temporal graph. These findings demonstrate that the model effectively addresses the three conflicting demands of the task: maintaining motion continuity, identifying infrequent fragmentation occurrences, and reducing false relationships between unrelated entities.


For the class-wise analysis and held-out evaluation reported here, 37 image pairs were used, with 25 image pairs used for training, 6 for validation, and 6 held out for final testing. The split is performed at the image-pair level, such that all object-pair samples from a given consecutive-frame pair remain within the same subset. The resulting class distributions are given in Table~\ref{tab:class_dist}.
 
\begin{table*}[h]
    \centering
    \caption{Class-wise sample counts of the Stage-2 dataset for the training, validation, and held-out test splits.}
    \label{tab:class_dist}
    \small
    \begin{tabular}{lccccc}
        \toprule
        Dataset split & Image pairs & MOVE (0) & BREAKUP (1) & NONE ($-1$) & Total object pairs \\
        \midrule
        Training      & 25 & 1116 & 45 & 5141 & 6302 \\
        Validation    & 6  & 199  & 15 & 1221 & 1435 \\
        Held-out test & 6  & 292  & 11 & 1718 & 2021 \\
        \midrule
        Total         & 37 & 1607 & 71 & 8080 & 9758 \\
        \bottomrule
    \end{tabular}
\end{table*}
 
The NONE class therefore strongly dominates the dataset, while BREAKUP constitutes a relatively small fraction of the samples. Thus, the reported perfect BREAKUP recall of 1.000 corresponds to correctly identifying all 11 BREAKUP samples in the held-out test set.
The spatial gating thresholds directly influence this performance. By limiting candidate connections to a centroid displacement of no more than 64 pixels or a normalized distance of 8.0, the inference method eliminates numerous physically implausible pairings prior to categorization. This alleviates the continuous difficulty of candidate matching and directs the classifier toward locally plausible relationships. In dense spray zones, where several droplets and ligaments may be in close proximity, such gating is crucial for managing ambiguity. The observed test performance should be seen as the result of a hybrid method that integrates learned relational inference with physically motivated candidate removal.
\subsubsection{Feature correlation analysis}
\begin{figure}[h!]
    \centering
    \includegraphics[width=\linewidth]{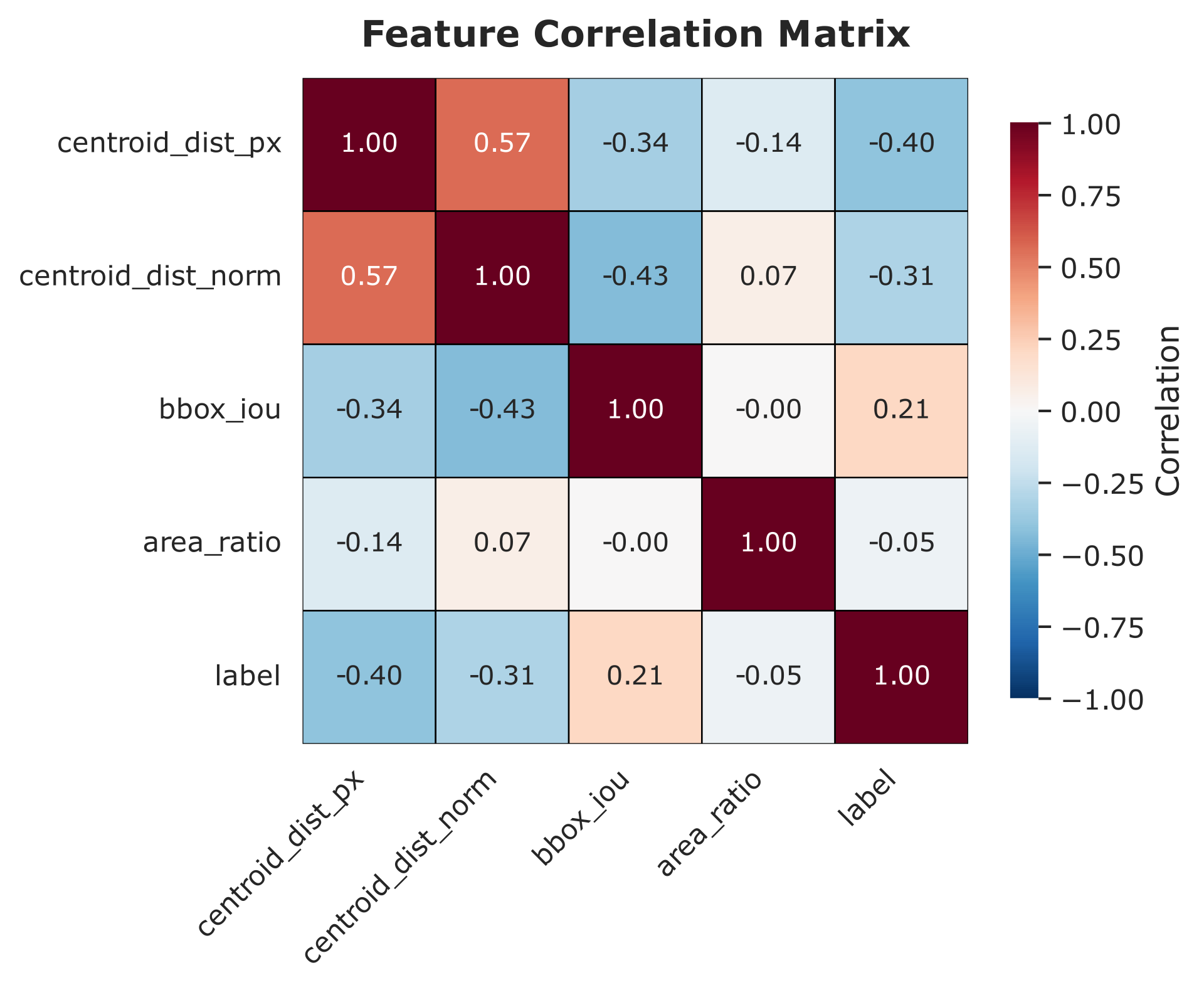}
    \caption{Correlation matrix for the five geometric input features together with the class label. Off-diagonal feature correlations are moderate, with the largest absolute correlation observed between centroid distance and normalized centroid distance, confirming that the selected features are complementary rather than redundant. Near-zero correlations involving \texttt{area\_ratio} indicate that size evolution provides information largely independent of spatial displacement and overlap cues.}
    \label{fig:feature_correlation}
\end{figure}

Fig.~\ref{fig:feature_correlation} displays the correlation matrix for the four geometric input features alongside the class label, resulting in a total of five investigated variables. The observed correlations are predominantly moderate, with the highest absolute correlation recorded at $|r| = 0.57$ between \texttt{centroid\_dist\_px} and \texttt{centroid\_dist\_norm}. This is anticipated, as both variables measure displacement information at varying scales. Significantly, no pair of input features demonstrates near-complete redundancy, suggesting that the chosen feature set include complimentary rather than irrelevant physical information.

The moderate correlation between the two centroid-distance variables confirms that both encode related but not identical spatial information. One reflects the raw displacement magnitude in pixel space, while the other normalizes it relative to object scale, thereby providing a more physically interpretable measure across objects of different sizes. Similarly, \texttt{bbox\_iou} contributes overlap-based evidence that cannot be inferred solely from centroid proximity, especially when objects deform or elongate between frames.

A notable observation is the near-zero correlation between \texttt{area\_ratio} and the centroid-distance. This indicates that size evolution and spatial displacement behave as largely independent physical cues. Such independence is highly desirable for breakup analysis, because a true fragmentation event may involve strong area change even when centroid motion is small, while a pure continuation may involve large displacement with minimal change in projected area. The low redundancy among these features therefore justifies their joint use and helps explain why architectures capable of modeling feature interactions, especially TransformerMLP, outperform simpler baselines.

The inclusion of the class label in the correlation matrix is also useful at an exploratory level, because it provides a compact view of how each input variable relates to the target categories. While these pairwise correlations do not by themselves capture the nonlinear decision boundaries learned by the classifier, they still indicate that the selected variables carry discriminative information relevant to temporal relationship prediction. The correlation structure supports the design methodology of the feature set: each variable captures a distinct aspect of temporal object behavior, and their combined use provides the classifier with a physically grounded yet sufficiently diverse representation of motion, overlap, and shape evolution.
\subsubsection{Temporal visualization and lineage reconstruction}
Fig.~\ref{fig:two_stage_framework} presents the overall inference pipeline of the proposed two-stage framework. Following object detection, candidate object pairs from consecutive frames are represented using the selected geometric variables and classified as \textsc{MOVE}, \textsc{BREAKUP}, or \textsc{NONE}. These predicted temporal relationships are then used to reconstruct lineage across the image sequence. This representation is particularly important for spray breakup analysis because it explicitly supports one-to-many transitions, which cannot be handled naturally by conventional one-to-one tracking methods. As a result, the framework enables physically interpretable lineage reconstruction and supports downstream extraction of breakup-related temporal statistics. The current framework is trained and tested on a single test condition, and the results presented are currently limited in applicability to the test condition.

\begin{figure*}[h]
    \centering
    \includegraphics[width=0.98\textwidth]{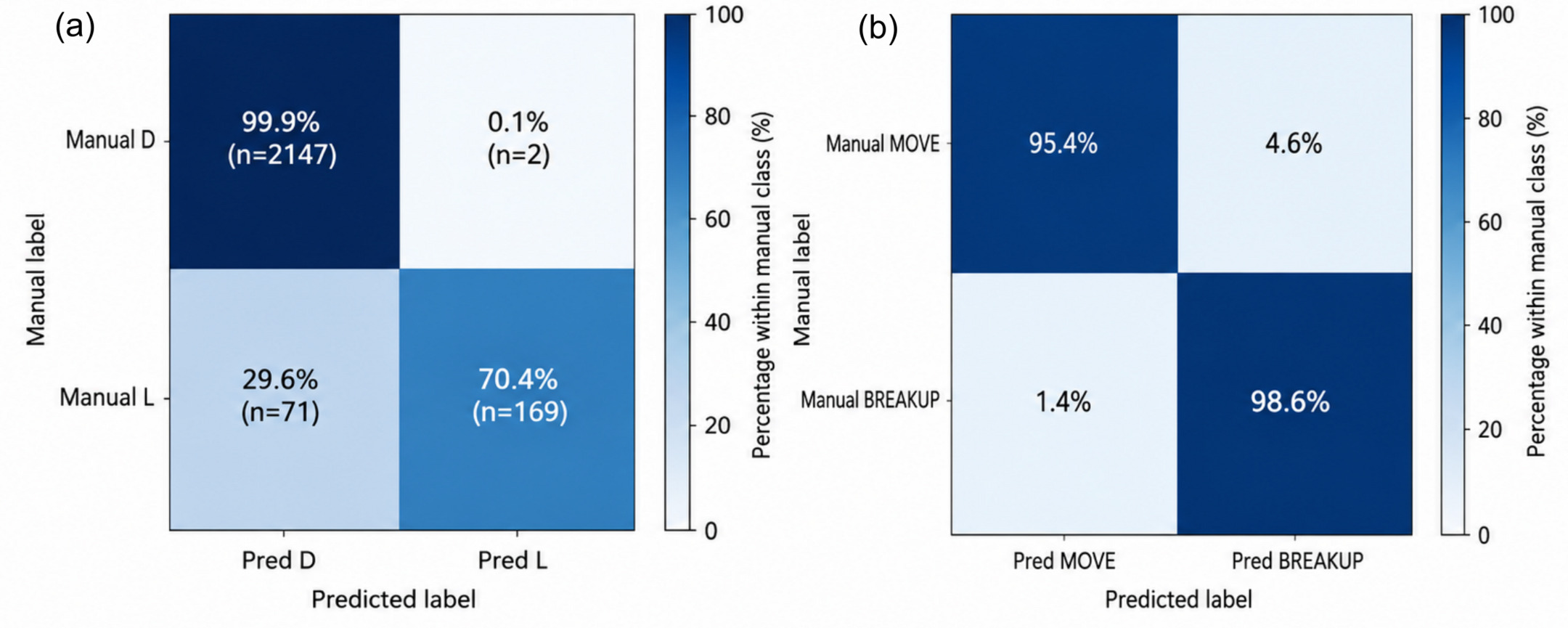}
    \caption{Confusion matrix for two stages. (a) Study-1 confusion matrix for Ligaments (L) and Droplets (D), and (b) for stage-2.}
    \label{fig:confusion}
\end{figure*}

\subsection{Confusion matrix}\label{subsec:confusion}
The classification performance of both stages of the proposed framework is evaluated using confusion matrices, shown in Fig.~\ref{fig:confusion}(a) and Fig.~\ref{fig:confusion}(b) for Stage 1 and Stage 2 respectively.
For Stage 1, the confusion matrix is reported for the single best performing configuration (Exp3, Nog = 250, Nsyn, selected by maximizing the product of overall accuracy and detection F1-score, a criterion applied identically across all experiments before inspecting the class-wise breakdown. This ensures the selected configuration was not chosen after the fact to present a more favorable result. The observed pattern, near perfect droplet recall (greater than 99.8\%) and ligament recall in the 65 to 70\% range with classification errors consistently one directional (ligament misclassified as droplet, not the reverse), is found across all tested configurations, confirming that the best case is representative of stable model behavior rather than an outlier. All ground truth annotations were produced through a single fixed manual segmentation protocol, so the confusion matrix reflects the model's ability to reproduce a consistently defined ground truth. The best case confusion matrix is therefore considered a faithful and sufficient summary of Stage 1 classification performance.
For Stage 2, as shown in Fig.~\ref{fig:confusion}(b), the proposed framework correctly classifies 95.4\% of manually labeled MOVE relationships, with only 4.6\% misidentified as BREAKUP. Similarly, 98.6\% of manually labeled BREAKUP relationships are correctly classified, with only 1.4\% misclassified as MOVE. The high values along the diagonal confirm the ability of the Stage 2 model to reliably distinguish object motion from fragmentation events, while the small off-diagonal values indicate minimal confusion between the two relationship classes, demonstrating the robustness of the relationship classification stage across the complete dataset.

Several limitations of the present study should be acknowledged. While Stage~1 was validated across six datasets spanning a range of Weber numbers and Carbopol concentrations within a single impinging-jet configuration, generalization to substantially different atomizer geometries, fluid systems beyond Carbopol gels, or different imaging setups remains to be established. Ligament recall in Stage~1, while consistent across configurations, is limited to approximately 70\%, primarily due to class imbalance and the inherent geometric ambiguity between compact ligaments and elongated droplets near the class boundary. Strategies such as class-weighted sampling, targeted synthetic generation of near-boundary ligament morphologies, or additional features may improve performance in this class. The temporal model in Stage~2 operates on pairwise features between consecutive frames and does not exploit longer-range temporal context.

\section{Conclusions}
\label{sec:conclusions}
This study presents a two-stage deep learning framework for the automated detection, temporal association, and lineage reconstruction of ligament breakup in high-speed shadowgraphy images of atomizing non-Newtonian liquid sheets. A liquid sheet breakup and droplet formation are fundamentally a topology-changing process in which a single parent structure may persist, deform, or fragment into multiple descendants over time. Such behavior is not naturally handled by conventional tracking methods based on one-to-one temporal assignment. To address this limitation, the proposed framework combines frame-wise object detection (study-I) with pairwise temporal relationship classification and facilitates breakup-aware lineage reconstruction (study-II).

In Study~I, a Faster R-CNN detector with a ResNet-50--FPN backbone is trained to detect and classify ligaments and droplets from cropped spray images (256 $\times$ 256 px). A morphology-preserving synthetic data generation strategy is employed to address annotation scarcity while maintaining realistic object morphology and spatial arrangement. The results demonstrated that synthetic augmentation is most effective when used to complement rather than replace original experimental images. The framework was evaluated across six experimental datasets spanning multiple Carbopol concentrations and Weber numbers, demonstrating consistent detection performance across meaningfully
different atomization regimes. The best configuration
($N_\text{og} = 250$, $N_\text{syn} = 250$) achieved a held-out test $F_1$-score of 0.872. Classification performance evaluated through the confusion matrix showed near-perfect droplet recall (99.9\%), confirming reliable separation of the dominant class. Ligament recall reached 70.4\%, with residual misclassifications occurring one-directionally (ligament predicted as droplet), attributable primarily to class imbalance and geometric overlap near the class boundary between compact ligaments and elongated droplets.

In Study~II, temporal reasoning is defined as a three-class relationship-prediction problem over candidate object pairs across consecutive frames. The relationship classifier used five physics-informed geometric features, namely centroid distance, normalized centroid distance, bounding-box intersection-over-union, area ratio, and label consistency. Feature correlation analysis confirmed that these inputs are largely complementary rather than redundant ($\max |r| = 0.57$), motivating the use of a self-attention mechanism capable of learning nonlinear feature interactions. Among the five architectures benchmarked, TransformerMLP produced the best and most consistent results across cross-validation folds, achieving an accuracy of 86.1\%, a precision of 93.2\%, and an $F_1$-score of 0.887 on the held-out test split. The confusion matrix for Stage~2 confirmed that 95.4\% of MOVE relationships and 98.6\% of BREAKUP relationships were correctly classified, with minimal off-diagonal confusion between the two classes. Critically, the model achieved perfect recall for the BREAKUP class, a property of direct physical importance because a missed fragmentation event would irreversibly disrupt the reconstructed parent--child lineage.

By combining the outputs of the two studies, the framework enables the construction of temporal graphs that explicitly encode continuation and fragmentation events across pairs. The proposed framework, therefore, extends spray image analysis beyond frame-wise detection and conventional tracking methods. It provides a physically interpretable and computationally effective representation of breakup dynamics.

The findings indicate that the fragmentation-dominated temporal analysis can be addressed effectively through relationship-aware learning based on reliable object detection. The proposed methodology reduces dependence on labor-intensive manual temporal annotation and offers a practical route toward automated analysis of breakup processes in complex multiphase flows. Future research should explore longer temporal context, graph-based or sequence-based temporal reasoning, instance-level representations, and validation across broader atomization regimes, conditions, imaging conditions, and fluid compositions. This study also provides future directions that can enable automatic reconstruction of fragmentation trees and the extraction of physically significant descriptors such as fragment multiplicity, parent--child inheritance, area-based consistency, and centroid-derived motion information.

\section*{CRediT authorship contribution statement}
\textbf{Vrushank Ahire:} Writing -- review \& editing, Writing -- original draft, Visualization, Software, Methodology, Formal analysis, Conceptualization. \textbf{Vivek Kurumanghat:} Writing -- review \& editing, Writing -- original draft, Visualization, Validation, Methodology, Formal analysis, Data curation, Conceptualization. \textbf{Mudasir Ganaie:} Writing -- review \& editing, Validation, Supervision, Software, Resources, Methodology, Investigation, Formal analysis, Conceptualization. \textbf{Lipika Kabiraj:} Writing -- review \& editing, Validation, Supervision, Resources, Project administration, Methodology, Investigation, Funding acquisition, Formal analysis, Data curation, Conceptualization.

\section*{Declaration of competing interest}
The authors declare the following financial interests/personal relationships which may be considered as potential competing interests: Lipika Kabiraj reports financial support was provided by Indian Space Research Organization. If there are other authors, they declare that they have no known competing financial interests or personal relationships that could have appeared to influence the work reported in this paper.

\section*{Acknowledgments}
We acknowledge financial support for research provided by the Indian Space Research Organization (Project Nr. ISRO/RES/3/782/18-19) and SERB DST, Govt. of India (Project Nr. SPG/2022/002031). We also acknowledge the Department of Computer Science and Engineering at IIT Ropar for providing the computational facility.

\appendix
\section{Supplementary data}
Supplementary material related to this article can be found online at \url{https://doi.org/10.1016/j.ijmultiphaseflow.2026.105974}.

\section*{Data availability}
Data will be made available on request.

\bibliographystyle{elsarticle-harv}             
\bibliography{references}

\end{document}